\documentclass{article}

\usepackage[preprint]{cpal_2026}

\usepackage{url}
\usepackage{graphicx}%
\usepackage{subcaption}
\usepackage{multirow}%
\usepackage{amsmath,amssymb,amsfonts}%
\usepackage{amsthm}%
\usepackage{mathrsfs}%
\usepackage[title]{appendix}%
\usepackage{xcolor}%
\usepackage{textcomp}%
\usepackage{manyfoot}%
\usepackage{booktabs}%
\usepackage{mathtools}
\usepackage{listings}%
\usepackage{upgreek}
\usepackage{wrapfig}

\theoremstyle{thmstyleone}%
\newtheorem{theorem}{Theorem}
\newtheorem{proposition}[theorem]{Proposition}%

\theoremstyle{thmstyletwo}%
\newtheorem{remark}{Remark}%

\theoremstyle{thmstylethree}%

\title{Deep Linear Discriminant Analysis Revisited}

\author{%
  Maxat Tezekbayev\textsuperscript{1}, ~Rustem Takhanov\textsuperscript{1}, ~Arman Bolatov\textsuperscript{2,4}, ~Zhenisbek Assylbekov\textsuperscript{3,4}\thanks{Corresponding author.} \\
  \textsuperscript{1}Department of Mathematics, Nazarbayev University, Astana, Kazakhstan \\
  \textsuperscript{2}Machine Learning Department, Mohamed bin Zayed University of Artificial Intelligence, Abu Dhabi, UAE \\
  \textsuperscript{3}Department of Mathematical Sciences, Purdue University Fort Wayne, Fort Wayne, IN, USA \\
  \textsuperscript{4}Remote Sensing Department, National Center of Space Research and Technology, Almaty, Kazakhstan \\
  \texttt{maxat.tezekbayev@alumni.nu.edu.kz, rustem.takhanov@nu.edu.kz, arman.bolatov@mbzuai.ac.ae, zassylbe@pfw.edu}%
}

\begin{document}

\maketitle

\begin{abstract}
We show that for unconstrained Deep Linear Discriminant Analysis (LDA) classifiers,
maximum-likelihood training admits pathological solutions in which class means drift together, covariances collapse,
and the learned representation becomes almost non-discriminative.
Conversely, cross-entropy training yields excellent accuracy but decouples the head from the underlying generative model,
leading to highly inconsistent parameter estimates.
To reconcile generative structure with discriminative performance, we introduce the
\emph{Discriminative Negative Log-Likelihood} (DNLL) loss, which augments the LDA log-likelihood with a simple penalty on the mixture density.
DNLL can be interpreted as standard LDA NLL plus a term that explicitly discourages regions where several classes are simultaneously likely.
Deep LDA trained with DNLL produces clean, well-separated latent spaces, matches the test accuracy of softmax classifiers on synthetic data and standard image benchmarks,
and yields substantially better calibrated predictive probabilities, restoring a coherent probabilistic interpretation to deep discriminant models.
\end{abstract}

\section{Introduction}

Classical Linear Discriminant Analysis (LDA) dates back to the pioneering works of \citet{fisher1936use} and \citet{rao1948}. It is one of the simplest and most widely used classification methods, with a clear probabilistic interpretation. Let \(X \in \mathbb{R}^d\) be an input vector and \(Y \in \{1,2,\dots,C\}\) the corresponding class label. LDA assumes the following generative model:
\begin{itemize}
  \item The labels follow a categorical distribution with probabilities $\pi=(\pi_1,\ldots,\pi_C)$\footnote{To avoid confusion with the constant $\uppi$, we reserve $\pi$ (and $\pi_c$) for class prior probabilities, while $\uppi$ denotes the mathematical constant.
} on the probability simplex $\Delta^{C-1}$:
  \begin{equation}
    \Pr[Y=c] = \pi_c,\qquad \pi_c \ge 0,\ \sum_{c=1}^C \pi_c = 1,
    \label{eq:priors}
  \end{equation}
  \item The class-conditional distribution is Gaussian:
  \begin{equation}
    X \mid Y = c\,\, \sim\,\, \mathcal{N}(\mu_c, \Sigma),
    \label{eq:class_cond}
  \end{equation}
  where the covariance matrix \(\Sigma\) is the same for all classes.
\end{itemize}

Specifying the marginal distribution of \(Y\) and the conditional distribution of \(X~\mid~Y\) in this way defines a joint distribution over \((X,Y)\). The model parameters\footnote{Throughout, $\theta$ denotes the collection of all learnable parameters of the model under consideration;
its exact contents will be clear from context.} \(\theta:=(\pi, \{\mu_c\}_{c=1}^C, \Sigma)\) can therefore be estimated by maximum likelihood. Once the model is fitted, prediction for a new input \(x\) is obtained by
\[
  \hat y(x) \;=\; \arg\max_{c} p_{\hat\theta}(x, Y=c),
\]
where \(p_{\hat{\theta}}(x,y)\) is the learned joint density of \((X,Y)\); equivalently, this is \(\arg\max_c p_{\hat\theta}(Y=c \mid X=x)\).

It is often convenient to work with the corresponding discriminant functions
\(\{\delta_c(x)\}_{c=1}^C\), defined by
\begin{equation}
  \delta_c(x)
  \;:=\;
  \log \pi_c
  - \frac{1}{2}\log\left(\det\Sigma\right)
  - \tfrac{1}{2}(x-\mu_c)^\top \Sigma^{-1}(x-\mu_c).
  \label{eq:lda-discriminants}
\end{equation}
The classifier can then be written as \(\hat y(x) = \arg\max_c \hat\delta_c(x)\).

A major limitation of classical LDA is that it yields linear decision boundaries.
This can be too restrictive when the classes are not linearly separable in the input space. Several approaches have been proposed to circumvent this limitation. Kernel methods, such as Kernel Fisher Discriminant Analysis \citep{mika1999fisher}, implicitly map the data into a high-dimensional feature space where linear separation is more plausible. Another approach is to model each class not by a single Gaussian but by a Gaussian mixture, leading to Mixture Discriminant Analysis \citep{hastie1996discriminant}.

\begin{wrapfigure}{r}{0.45\textwidth} 
    \centering
    \includegraphics[width=0.43\textwidth]{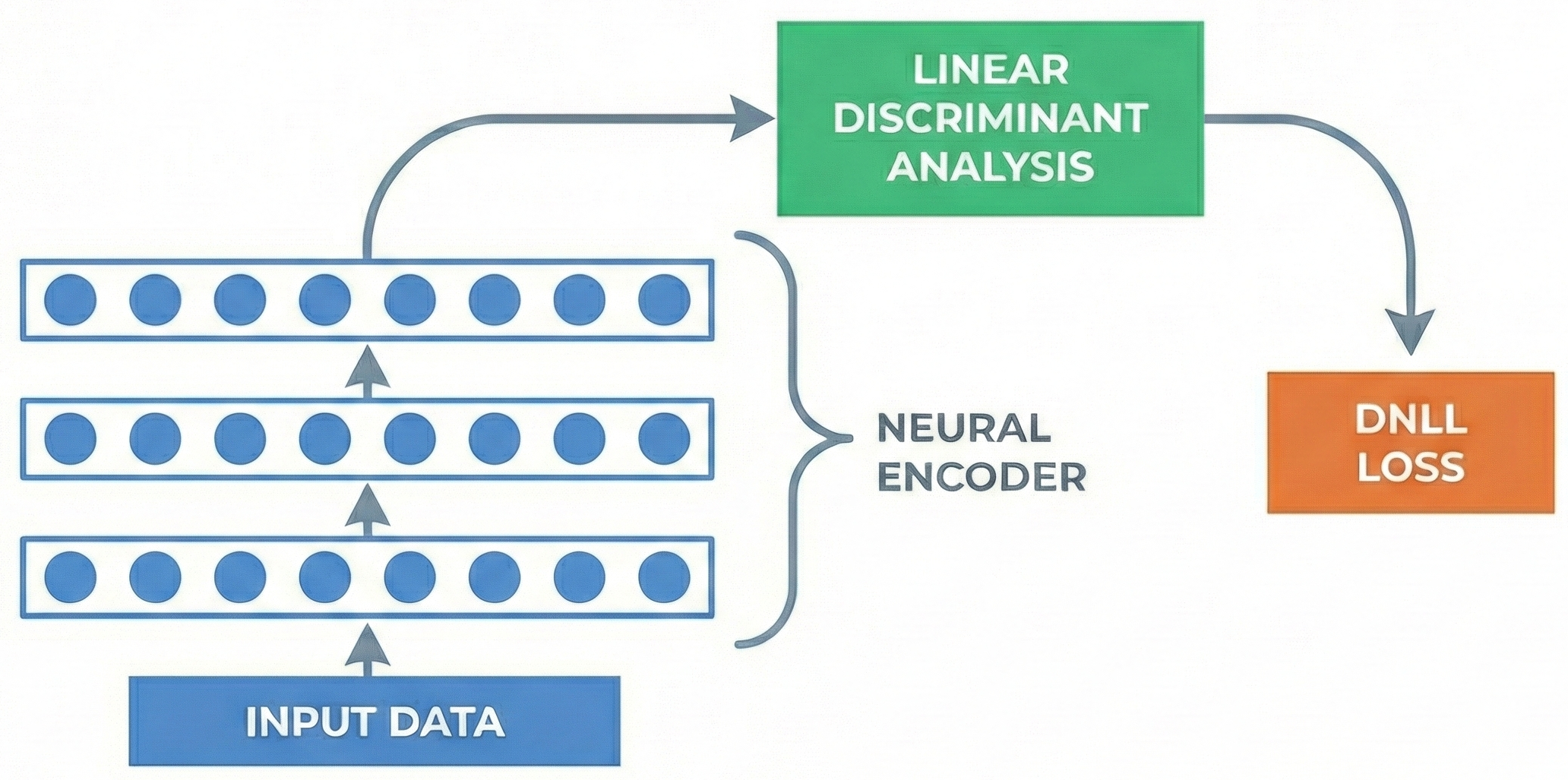}
    
    \caption{Deep LDA architecture. A neural encoder maps input data to a latent space where LDA is applied, and the network is optimized using the proposed Discriminative Negative Log-Likelihood (DNLL) loss.}
    \label{fig:dnll_arch}
\end{wrapfigure}

In this paper we focus on \emph{Deep LDA} \citep{DBLP:journals/corr/DorferKW15}, which feeds input into a neural encoder and applies an LDA head in the resulting latent space (Figure~\ref{fig:dnll_arch}).
\citet{DBLP:journals/corr/DorferKW15} do not optimize the likelihood of the LDA model; instead, they attempt to maximize Fisher’s generalized eigenvalue objective. Directly optimizing the classical Fisher objective turned out to be problematic, and they had to resort to a heuristic that emphasizes eigen-directions with low discriminative power. As a result, their final training objective departs from the original Fisher criterion.

At the same time, LDA is fundamentally a probabilistic model, and it would be highly desirable to train its deep variant by maximum likelihood, thereby retaining a coherent generative interpretation. However, as we show in Section~2, when no constraints are imposed on either the encoder or the LDA head, maximum-likelihood training of Deep LDA can become almost completely decoupled from discrimination. The likelihood is maximized by pathological solutions in which embeddings collapse tightly around their class means, the shared covariance matrix becomes nearly singular, and different class means drift closer together. Such configurations achieve high likelihood while yielding heavily overlapping clusters, poor classification accuracy, and unreliable confidence estimates.

Switching from the generative paradigm to a discriminative one---by treating the LDA discriminant functions \eqref{eq:lda-discriminants} as logits and optimizing cross-entropy---does improve classification accuracy. Yet, in this regime the LDA parameters are no longer estimated consistently as those of the underlying generative model, a phenomenon that can already be observed in classical LDA.

This leads to the central question of the present work:
\begin{quote}
\centering
\emph{Can we train Deep LDA by maximum likelihood without sacrificing either parameter consistency or discriminative power?}
\end{quote}

In Section~\ref{sec:dnll} we address this question by introducing a penalized likelihood objective, which we call the \textbf{Discriminative Negative Log-Likelihood (DNLL)}. Concretely, we augment the LDA log-likelihood with a penalty term that explicitly promotes class separation: it prevents degenerate solutions from increasing the likelihood without bound and encourages the encoder to map inputs into well-separated clusters around the class means. Our analysis on synthetic data further shows that, for well-separated classes, optimizing this penalized loss yields parameter estimates that remain close to the maximum-likelihood solution.

In Section~\ref{sec:real-data} we train Deep LDA with DNLL loss on image classification tasks. The resulting model achieves test accuracy comparable to a model with the softmax head trained with cross-entropy. Crucially, it also yields substantially better calibrated predictive probabilities, as quantified by reliability diagrams and Expected Calibration Error (ECE), supporting the view that likelihood-based training with an explicit generative head can improve the quality of confidence estimates without compromising accuracy. Code to reproduce our experiments is available at \url{https://github.com/zh3nis/DNLL}.

To our knowledge, no prior work has analyzed the failure modes of Deep LDA under maximum likelihood or proposed a principled, likelihood-based training that enforces discriminative separation while preserving the generative structure of LDA.

\section{Maximum-Likelihood Training: A First Look}\label{sec:mle}
Before turning to deep architectures, we first check that classical (non-deep) LDA can indeed be fitted accurately by gradient-based maximum-likelihood on a simple synthetic dataset.

\subsection{Warm-up: Classical LDA}

We work with the generative model in \eqref{eq:priors}--\eqref{eq:class_cond}. 
 Given a labeled sample
\(\{(x_1,y_1),\dots,(x_n,y_n)\}\) drawn from \eqref{eq:priors}--\eqref{eq:class_cond}, the negative log-likelihood is
\begin{equation}
\label{eq:avg-loglike}
\mathcal{L}_{\text{NLL}}(\theta)
=-\frac{1}{n}\sum_{i=1}^n \log\!\Big( \pi_{y_i}\,\phi(x_i;\,\mu_{y_i},\Sigma) \Big),
\end{equation}
where, for \(x\in\mathbb{R}^d\),
\[
\phi(x;\mu,\Sigma)
=(2\uppi)^{-d/2}\,(\det\Sigma)^{-1/2}
\exp\!\left\{-\tfrac{1}{2}(x-\mu)^\top\Sigma^{-1}(x-\mu)\right\}
\]
is the density of \(\mathcal{N}(\mu,\Sigma)\).

\paragraph*{Optimization}
We minimize \(\mathcal{L}_{\text{NLL}}(\theta)\) using Adam~\citep{DBLP:journals/corr/KingmaB14}. To keep the optimizer unconstrained, we use the following reparameterizations:
(i) logits \(\alpha\in\mathbb{R}^C\) with \(\pi=\mathrm{softmax}(\alpha)\), and
(ii) a Cholesky factorization \(\Sigma = LL^\top\), where \(L\) is lower triangular with a positive diagonal. This guarantees \(\pi\in\Delta^{C-1}\) and \(\Sigma \succ 0\) while allowing standard automatic differentiation.

\begin{figure}[htbp]
  \centering
  \includegraphics[width=.65\textwidth]{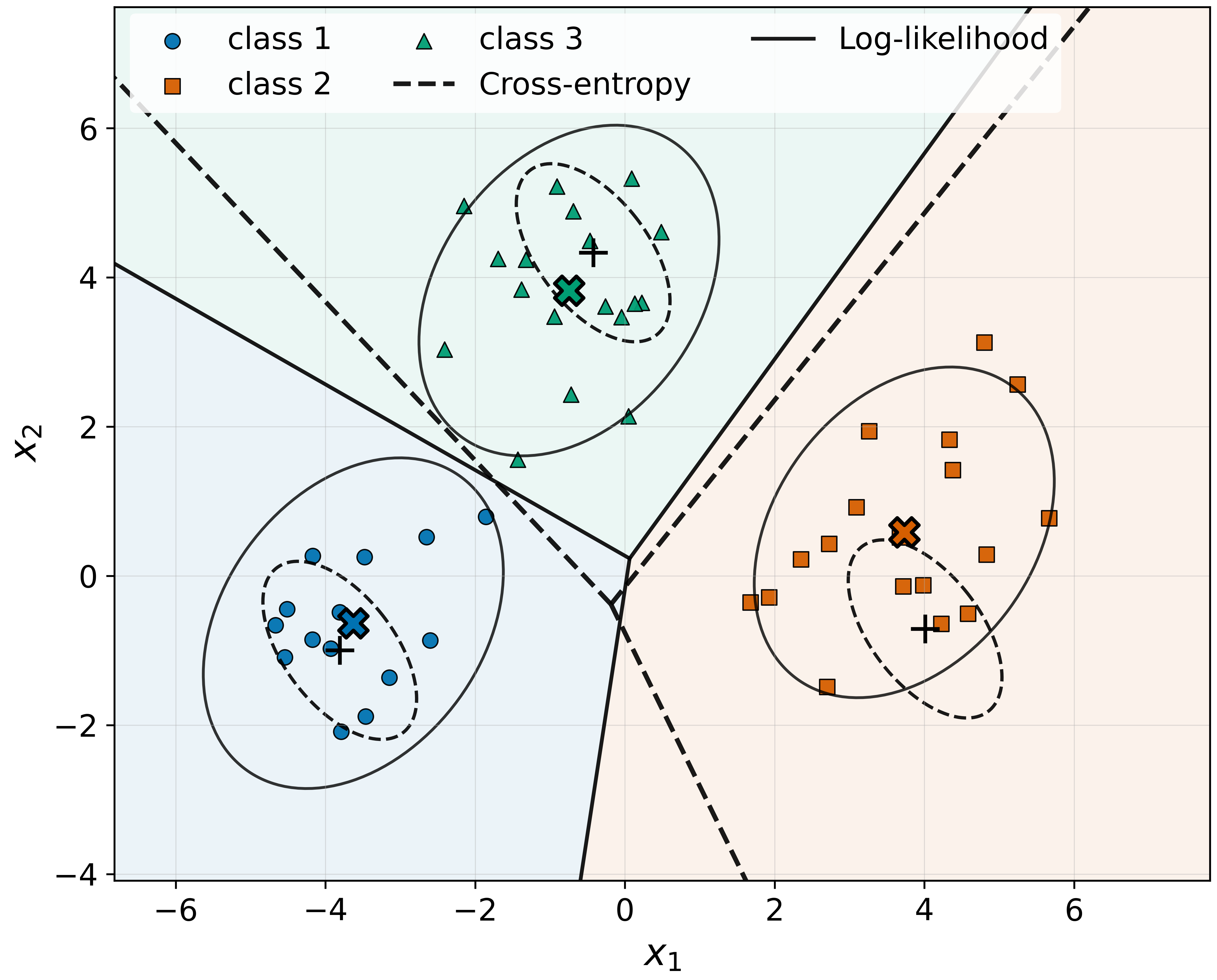}
  \caption{Classical three-class LDA fitted on synthetic data by maximum likelihood (solid ellipses, solid decision boundaries, means marked by `$\times$') and by minimizing cross-entropy with discriminant scores used as logits (dashed ellipses, dashed boundaries, means marked by `$+$'). The  Gaussian components (means and covariances) differ noticeably, illustrating that cross-entropy training does not recover the maximum-likelihood parameters of the generative LDA model. Ellipses show the 90\% contours of the learned Gaussian classes.}
  \label{fig:lda_ce}
\end{figure}

\paragraph*{Result}
On data generated from \eqref{eq:priors}--\eqref{eq:class_cond}, gradient-based optimization reliably recovers the maximum-likelihood solution; see Figure~\ref{fig:lda_ce}.
In particular, the learned priors, means, and covariance closely match the ground-truth parameters, and the decision boundaries coincide with those obtained from the closed-form LDA estimates. This confirms that the classical model can be trained by gradient-based likelihood maximization.

\paragraph*{Cross-entropy training and inconsistency}
Alternatively, we can discard the generative interpretation and train the LDA head
\emph{discriminatively}, by treating the discriminant functions \eqref{eq:lda-discriminants}
as logits. Indeed, under the LDA model, Bayes' rule gives
\begin{align}
p_\theta(Y=c \mid x)
&=
\frac{\pi_c\,\varphi(x;\mu_c,\Sigma)}{\sum_{j=1}^C \pi_j\,\varphi(x;\mu_j,\Sigma)}
=
\frac{\exp(\delta_c(x))}{\sum_{j=1}^C \exp(\delta_j(x))}
\;=\;
\mathrm{softmax}\big(\delta(x)\big)_c,
\label{eq:lda-posterior-softmax}
\end{align}
where $\delta(x)=(\delta_1(x),\dots,\delta_C(x))^\top$ and the equality holds because
$\delta_c(x)=\log \pi_c+\log \varphi(x;\mu_c,\Sigma)$ up to an additive constant independent
of $c$ (which cancels in the normalization).

This suggests optimizing $\theta$ by minimizing the cross-entropy (negative conditional
log-likelihood) with logits $\delta_c(x)$:
\begin{align}
\ell_{\mathrm{CE}}(x,y;\theta)
&:= -\log p_\theta(y\mid x)
= -\log\frac{\exp(\delta_y(x))}{\sum_{c=1}^C \exp(\delta_c(x))}\notag\\
&= -\delta_y(x) + \log\sum_{c=1}^C \exp(\delta_c(x)),
\label{eq:lda-ce-loss}
\\
\mathcal{L}_{\mathrm{CE}}(\theta)
&:= \frac{1}{n}\sum_{i=1}^n \ell_{\mathrm{CE}}(x_i,y_i;\theta).
\label{eq:lda-ce-objective}
\end{align}
On the same synthetic data as above, we compare this procedure with gradient-based MLE.
The two objectives yield perfect training accuracies, but the estimated Gaussian components
differ substantially; see Figure~\ref{fig:lda_ce}. In particular, the cross-entropy solution
shifts the class means and shrinks the covariances, producing tighter ellipses that no longer
match the data-generating model. Thus, while cross-entropy training gives a good classifier,
it does \emph{not} recover the ground-truth parameters of the underlying LDA model and is
therefore an inconsistent estimator in this generative setting.

\subsection{A Cautionary Experiment: Likelihood Training of Deep LDA}

We now move to the deep case. For the Deep LDA experiment, we again generate a dataset 
\(\{(x_1,y_1),\ldots,(x_n,y_n)\}\) 
from the model in \eqref{eq:priors}--\eqref{eq:class_cond}, but pass inputs through a neural encoder. Specifically, we take \(f_\psi(\cdot)\) to be a two-layer feed-forward network with ReLU activations and 32 hidden units. The resulting LDA negative log-likelihood is
\begin{equation}
\label{eq:deep-lda-like}
\mathcal{L}_{\text{NLL}}(\theta)
=-\frac{1}{n}\sum_{i=1}^n
\log\!\left(
\pi_{y_i}\,
\phi\!\left(f_\psi(x_i);\ \mu_{y_i},\,\Sigma\right)
\right),
\end{equation}
with parameters \(\theta := (\psi, \pi, \{\mu_c\}_{c=1}^C, \Sigma)\).

\begin{figure}[htbp]
  \begin{minipage}[t]{.47\textwidth}
  \centering
  \includegraphics[width=\textwidth]{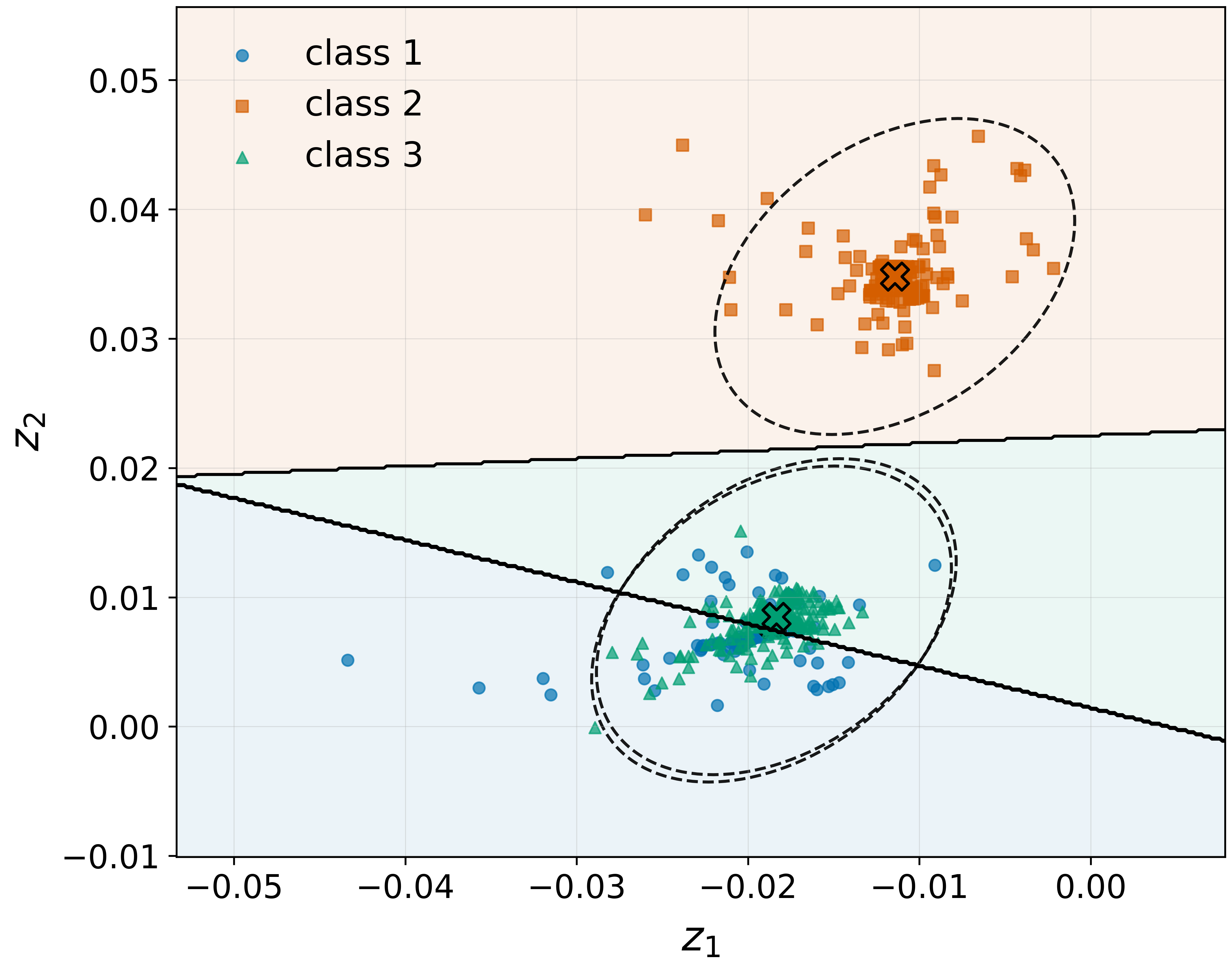}
  \caption{Deep LDA embeddings \(z_i=f_\psi(x_i)\) after likelihood training. 
  Two classes merge into a single cluster; samples concentrate tightly around their class centroids \(\{\mu_c\}\) while the shared covariance \(\Sigma\) becomes nearly singular ($|\Sigma|\approx3\cdot10^{-10}$). 
  Training accuracy: 67.2\%; test accuracy: 67.7\%.}
  \label{fig:deep_lda_emb}
  \end{minipage}
  \hfill
  \begin{minipage}[t]{.47\textwidth}
  \centering
  \includegraphics[width=\linewidth]{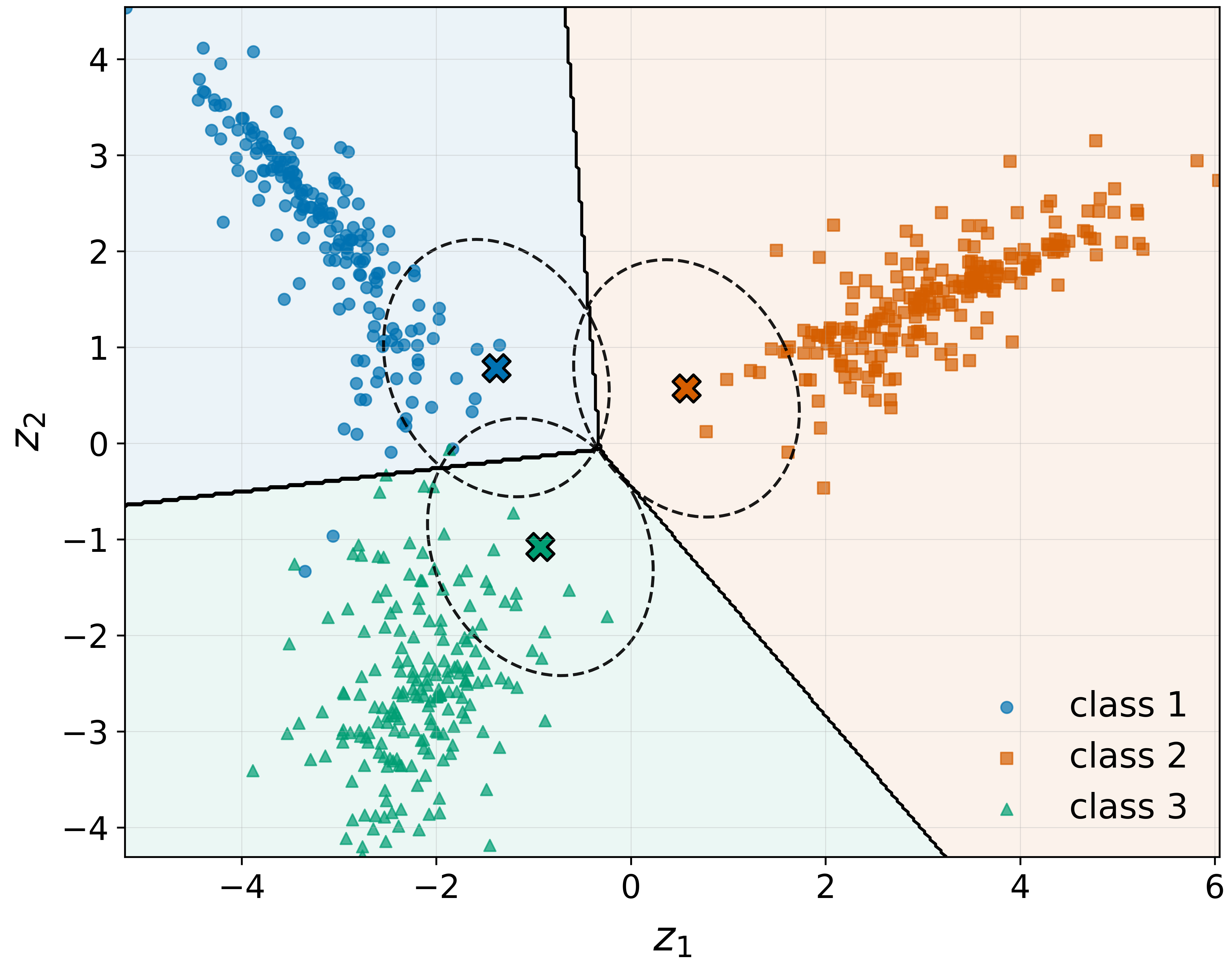}
  \caption{Deep LDA embeddings \(f_\psi(x_i)\) after cross-entropy training. 
  Gaussian components are located near the intersection of the boundaries rather than around the empirical class clouds yielding inconsistent parameter estimates. 
  Training accuracy: 99.4\%; test accuracy: 99.6\%.}
  \label{fig:deep_lda_ce}    
  \end{minipage}
\end{figure}

We draw \(20{,}000\) training and \(4{,}000\) test points. 
Training is run for 100 epochs using Adam with PyTorch’s default hyperparameters and a minibatch size of 256 (batch size 1024 at evaluation). The learned embeddings \(z_i = f_\psi(x_i)\) are shown in Figure~\ref{fig:deep_lda_emb}.

The failure mode is apparent: the optimizer does not enforce separation of all classes—two of the class clusters collapse into one. Moreover, for each labeled pair \((x, y)\), the encoder drives \(f_\psi(x)\) extremely close to its class mean \(\mu_y\), pushing the squared Mahalanobis distance
\[
\left(f_\psi(x)-\mu_{y}\right)^\top
\Sigma^{-1}
\left(f_\psi(x)-\mu_{y}\right)
\]
toward zero while simultaneously shrinking \(\Sigma\) toward near-singularity.
Both effects increase the Gaussian term 
\(\phi\big(f_\psi(x);\mu_{y},\Sigma\big)\), so the log-likelihood is maximized by highly concentrated, almost overlapping clusters.

Despite the very high likelihood, the classifier performs poorly: the final training and test accuracies are only 67.2\% and 67.7\%, respectively. This illustrates a fundamental misalignment between the unconstrained maximum-likelihood objective for Deep LDA and the goal of learning a well-separated, discriminative representation.

\subsection{Deep LDA with Cross-Entropy Loss}

The experiment above shows that unconstrained maximum-likelihood training of Deep LDA leads to a highly
degenerate solution. A natural alternative is to train the same
architecture \emph{discriminatively}. Concretely, we keep the encoder \(f_\psi\) and the LDA head, but instead of maximizing the log-likelihood,
we treat the discriminant scores
$\delta_c(f_\psi(x))$ as logits and minimize the cross-entropy loss.

We repeat the synthetic experiment from the previous subsection with the same data-generating model,
encoder architecture, optimizer, and training schedule, changing only the objective from
negative log-likelihood to cross-entropy. The resulting embeddings \(z_i=f_\psi(x_i)\) and decision boundaries
are shown in Figure~\ref{fig:deep_lda_ce}. In contrast to the likelihood-trained model, the three classes form distinct clusters. Train and test accuracies are both near perfect, confirming
that cross-entropy training yields an excellent classifier.

However, the parameters of the LDA head no longer align with the empirical class structure. The learned
means \(\{\mu_c\}\) sit near the intersection of the decision boundaries, and the dashed ellipses in Figure~\ref{fig:deep_lda_ce}---that are supposed to reflect the shared covaraince structure---have little resemblance
to the elongated class clusters produced by the encoder. In other words, the discriminative
training objective uses the LDA head merely as a convenient parametric form for logits, and the resulting
\((\pi,\{\mu_c\},\Sigma)\) are not close to the maximum-likelihood parameters of the underlying Gaussian
mixture. This mirrors the inconsistency observed in the shallow setting and illustrates that, for Deep
LDA as well, cross-entropy training destroys the probabilistic interpretation of the model parameters
even when classification performance is excellent.

\section{Discriminative Negative Log-Likelihood}
\label{sec:dnll}
Motivated by the failure modes in Section~2, we introduce a simple loss that preserves likelihood-based training while explicitly encouraging class separation.

\subsection{Definition}

Let $z = f_\psi(x)$ be the embedding of an input $x$ and
$\{\delta_c(z)\}_{c=1}^C$ the corresponding discriminant functions of
the LDA head in the latent space.
For a labeled example $(x,y)$, the classical negative log-likelihood
(NLL) of the generative LDA model is, up to an additive constant,
\begin{equation}
  \ell_{\text{NLL}}(z,y)
  \;=\;
  - \delta_y\big(z\big).
  \label{eq:lda-nll}
\end{equation}

To promote discrimination, we augment \eqref{eq:lda-nll} with a penalty
on the {unnormalized} class densities,
\begin{equation}
  R(z)
  \;=\;
  \sum_{c=1}^C \exp\bigl(\delta_c(z)\bigr).
  \label{eq:dnll-penalty}
\end{equation}
The proposed \emph{Discriminative Negative Log-Likelihood} (DNLL) is
defined as
\begin{align}
  \ell_{\text{DNLL}}(z,y;\theta)
  &:=
  - \delta_y\big(z\big)
  + \lambda
    \sum_{c=1}^C \exp\bigl(\delta_c(z)\bigr),
  \qquad \lambda \ge 0,\label{eq:dnll}\\
  \mathcal{L}_{\text{DNLL}}(\theta)&:=\frac1n\sum_{i=1}^n\ell_{\text{DNLL}}(f_\psi(x_i),y_i;\theta)
  \label{eq:dnll_full}
\end{align}
with $\lambda$ controlling the strength of the discriminative
regularization.  When $\lambda=0$ we recover pure maximum-likelihood
training, while $\lambda>0$ penalizes embeddings that lie in regions
where several class densities are simultaneously large.

For LDA we can rewrite \eqref{eq:dnll} in terms of Gaussian densities.
Since
\[
  \exp\!\bigl(\delta_c(z)\bigr)
  \;=\;
  \pi_c \, \phi(z; \mu_c, \Sigma)
  \,\cdot\,
  (2\uppi)^{d/2},
\]
the penalty \eqref{eq:dnll-penalty} is, up to a global constant, the
mixture density at $z$:
\[
  R(z)
  \;\propto\;
  \sum_{c=1}^C \pi_c \, \phi\bigl(z; \mu_c, \Sigma\bigr).
\]
Thus DNLL decomposes as
\begin{equation}
  \ell_{\text{DNLL}}(z,y)
  \;=\;
  \underbrace{-\log \left(\pi_y
  \phi\bigl(z; \mu_y, \Sigma\bigr)\right)}_{\text{NLL}}
  \;+\;
  \lambda\,
  \underbrace{\sum_{c=1}^C \pi_c \, \phi\bigl(z; \mu_c, \Sigma\bigr)}_{\text{mixture density}}
  \;+\; \text{const}.
  \label{eq:dnll-decomp}
\end{equation}
The first term is exactly the negative log-likelihood of the LDA model,
while the second term penalizes points that lie in regions of high
overall mixture density.  Intuitively, the optimizer is encouraged to
map each sample into an area where its own class density is high but
the competing class densities are low,
reducing overlap between the Gaussian components.

\paragraph{Why the density penalty separates classes.}
With an LDA head, the marginal embedding density is a shared-covariance Gaussian mixture
\(
p_\theta(z)=\sum_{c=1}^C \pi_c\,\phi(z;\mu_c,\Sigma).
\)
Up to a global constant, the DNLL regularizer equals this mixture density evaluated at the embedding,
\(
R(z)\propto p_\theta(z)
\)
(see \eqref{eq:dnll-decomp}), so DNLL penalizes training embeddings that fall in regions where the
\emph{overall} model density is large.

More precisely, writing the penalty term in the empirical DNLL objective gives
\begin{equation}
\label{eq:dnll-penalty-empirical}
\frac{\lambda}{n}\sum_{i=1}^n R(z_i)
\;\propto\;
\lambda\,\widehat{\mathbb E}\big[p_\theta(Z)\big]
\;:=\;
\frac{\lambda}{n}\sum_{i=1}^n p_\theta(z_i),
\end{equation}
i.e., DNLL minimizes the \emph{average mixture density evaluated on the training embeddings}.
This quantity is large when many training points lie in regions where multiple mixture components
simultaneously assign non-negligible probability mass (high overlap), so decreasing it discourages
embeddings from occupying such high-overlap areas and thus promotes separation.

A convenient \emph{intrinsic} (distribution-level) proxy for overlap in mixtures is the
\emph{information potential}
\(
C(p_\theta)=\int [p_\theta(z)]^2\,dz
=
\mathbb E_{Z\sim p_\theta}\big[p_\theta(Z)\big],
\)
which measures the self-overlap of the marginal density~\citep{ACU2021125578}.
Thus, the DNLL penalty $\widehat{\mathbb E}[p_\theta(Z)]$ can be viewed as a data-driven analogue of
$C(p_\theta)$: if the embedding marginal $p(z)$ induced by the encoder is close to the model mixture
$p_\theta(z)$ (as encouraged by the likelihood term), then
\(
\mathbb E_{Z\sim p}[p_\theta(Z)] \approx \mathbb E_{Z\sim p_\theta}[p_\theta(Z)] = C(p_\theta).
\)
(A quantitative bound relating these two expectations appears in Appendix~\ref{app:dnll-detailed}.)

For shared-covariance Gaussian mixtures, \(C(p_\theta)\) has a closed form:
\begin{equation}
C(p_\theta)
=
\sum_{i,j=1}^C \pi_i\pi_j\int \phi(z;\mu_i,\Sigma)\,\phi(z;\mu_j,\Sigma)\,dz
=
\sum_{i,j=1}^C \pi_i\pi_j\,\phi(\mu_i-\mu_j;0,2\Sigma),
\label{eq:collision-phi-1}
\end{equation}
and, equivalently,
\begin{equation}
C(p_\theta)
=
(4{\uppi})^{-d/2}|\Sigma|^{-1/2}\sum_{i,j=1}^C \pi_i\pi_j
\exp\!\Big(-\tfrac14\|\mu_i-\mu_j\|_{\Sigma^{-1}}^2\Big).
\label{eq:collision-phi-2}
\end{equation}

The decomposition in \eqref{eq:collision-phi-2} makes the effect of overlap explicit.
The diagonal terms \(i=j\) contribute
\(
(4\uppi)^{-d/2}|\Sigma|^{-1/2}\sum_i \pi_i^2,
\)
which is independent of the means but diverges as \(\det\Sigma\) approaches zero.
Thus, any objective that correlates with \(C(p_\theta)\) creates a barrier against covariance collapse.
The off-diagonal terms \(i\neq j\) contribute
\(
(4\uppi)^{-d/2}|\Sigma|^{-1/2}\pi_i\pi_j\exp(-\|\mu_i-\mu_j\|_{\Sigma^{-1}}^2/4),
\)
which is largest when \(\mu_i\) and \(\mu_j\) are close and decays exponentially with their Mahalanobis
distance. Minimizing these terms therefore induces an explicit \emph{repulsion} between distinct class
centers, favoring configurations with well-separated means.

Finally, DNLL applies this overlap-reduction mechanism \emph{where the data live}.
For a training embedding \(z\) of class \(y\), the NLL term encourages \(z\) to lie in a high-density
region of its own component \(\phi(\cdot;\mu_y,\Sigma)\), while the penalty \(\lambda\,p_\theta(z)\)
discourages \(z\) from lying in regions where other components also have non-negligible density.
Geometrically, this steers embeddings away from decision-boundary regions and toward regions in which the correct class dominates,
reducing overlap between components without abandoning the generative structure.

A detailed derivation of \eqref{eq:collision-phi-1}--\eqref{eq:collision-phi-2} and additional discussion appear in Appendix~\ref{app:dnll-detailed}.

\subsection{Deep LDA Trained with DNLL on Synthetic Data}

We now revisit the synthetic Deep LDA setup from
Section~\ref{sec:mle}, replacing the negative log-likelihood objective
\eqref{eq:deep-lda-like} with the DNLL loss \eqref{eq:dnll}.  The
data-generating model, encoder architecture, optimizer, and training
schedule are kept identical: we use the same three-class Gaussian
mixture in the input space, a two-layer ReLU network with 32 hidden
units as the encoder $f_\psi$, and train on $20{,}000$ examples with
Adam for 100 epochs (batch size 256 for training and 1024 at
evaluation).  The regularization coefficient
is set to $\lambda=.01$.

\begin{figure}[htbp]
  \centering
  \includegraphics[width=.65\textwidth]{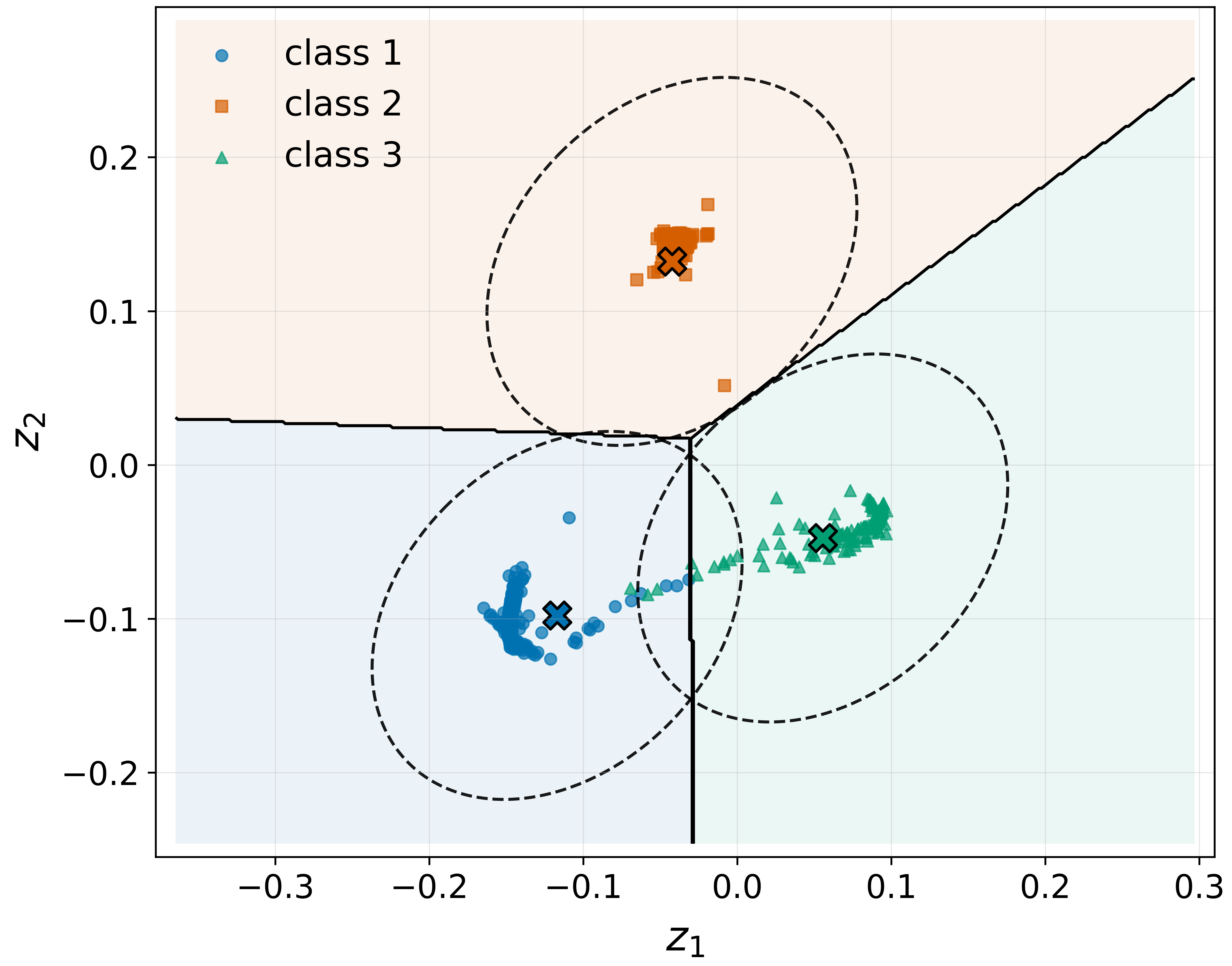}
  \caption{Deep LDA embeddings $z_i = f_\psi(x_i)$ after training with
  the proposed Discriminative NLL.   Unlike pure likelihood training
  (Figure~\ref{fig:deep_lda_emb}), DNLL yields well-separated clusters
  with Gaussian components aligned to the empirical class clouds, while
  achieving near-perfect train and test accuracies: 99.4\% and 99.5\% respectively.}
  \label{fig:deep_lda_dnll}
\end{figure}

Figure~\ref{fig:deep_lda_dnll}
shows the resulting embeddings
$z_i=f_\psi(x_i)$, the learned Gaussian components, and the induced
decision regions.  In sharp contrast to the pathological behaviour of
pure likelihood training (Figure~\ref{fig:deep_lda_emb}), the three
classes now form better-separated clusters in the latent space.  The
class means $\{\mu_c\}$ are located near the centers of the empirical
clouds.  At the same time, classification
performance matches that of cross-entropy training: both train and test
accuracies are above $99\%$ on this synthetic task.

These results suggest that DNLL successfully aligns the generative and
discriminative roles of the LDA head.  The likelihood term preserves
the probabilistic structure of the model and yields parameters close to
those of the underlying Gaussian mixture, while the density penalty
prevents the encoder from exploiting degenerate, non-discriminative
solutions.  In the remainder of the paper we further analyze this loss
by evaluating it on real image
classification benchmarks.

\section{Experiments on Image Classification}
\label{sec:real-data}

We now evaluate the Deep LDA model trained with the Discriminative
Negative Log-Likelihood of Section~\ref{sec:dnll} on image
classification benchmarks.  As in the synthetic experiments, the model
consists of an encoder $f_\psi$ followed by an LDA head.
The encoder and head parameters are trained jointly by minimizing the
empirical DNLL loss.  We compare this model to an identical encoder
equipped with a linear softmax head trained with cross-entropy.

A key difference from the synthetic setting is that, for image experiments, we impose
geometric constraints on the LDA head for stability and for a fair comparison to a linear
softmax layer. In our main configuration we use $d=C-1$ and a spherical covariance
$\Sigma=\sigma^2 I_d$. We also report an ablation over covariance parameterizations
(spherical, diagonal, and full) in Table~\ref{tab:image-results}.
Specifically, we use:
\begin{itemize}
\item embeddings in $\mathbb{R}^d$ with $d=C-1$, where $C$ is the number of classes;
\item a shared covariance matrix $\Sigma$ parameterized as either
(i) spherical $\sigma^2 I_d$ (main setting),
(ii) diagonal $\mathrm{diag}(\sigma_1^2,\ldots,\sigma_d^2)$, or
(iii) full $\Sigma\succ 0$ (Cholesky parameterization).
\end{itemize}

From the perspective of classical LDA, this is a natural choice:
for a $C$-class Gaussian model with shared covariance, the
between-class scatter matrix has rank at most $C-1$, and the Bayes
decision rule depends only on projections onto a $(C-1)$-dimensional
discriminant subspace.  Having the encoder produce embeddings directly
in $\mathbb{R}^{C-1}$ amounts to learning this discriminant subspace
end-to-end.  

The spherical covariance further simplifies the head and
keeps its number of trainable parameters on the same order as that of a
linear softmax layer: both have $\mathcal{O}(Cd)$ parameters when
$\mu_1,\dots,\mu_C\in\mathbb{R}^d$ and $d=C-1$. The spherical covariance keeps the head lightweight, whereas diagonal and full $\Sigma$
increase flexibility at the cost of additional parameters and potential overfitting.

With these constraints, the LDA discriminants in the latent space take
the form
\[
  \delta_c(z)
  \;=\;
  \log \pi_c - \frac{1}{2\sigma^2}\,\|z-\mu_c\|_2^2
  \quad\text{for } z = f_\psi(x)\in\mathbb{R}^{C-1},
\]
and the DNLL loss in \eqref{eq:dnll} reduces to a combination of the
class-conditional squared distances and the mixture density in
$\mathbb{R}^{C-1}$.

\paragraph*{Datasets}
We evaluate on three standard image classification datasets.
{Fashion-MNIST}~\citep{xiao2017fashion} contains 70\,000 grayscale
images of size $28\times 28$ from $C=10$ classes, split into 60\,000
training and 10\,000 test samples.
{CIFAR-10} and {CIFAR-100}~\citep{krizhevsky2009learning}
consist of $32\times 32$ color images with $C=10$ and $C=100$ classes,
respectively, each with 50\,000 training and 10\,000 test images.
Following common practice, we apply standard data augmentation to
CIFAR-10/100 (random horizontal flips and random crops with padding)
and use only per-pixel normalization on Fashion-MNIST.

\paragraph*{Encoder architecture and latent dimensionality}
All experiments share the same convolutional encoder, differing only in
the number of input channels.  The encoder consists of three
convolutional blocks with channel widths $\{64, 128, 256\}$.  Each block
applies two $3\times 3$ \mbox{$\mathrm{Conv}\!-\!\mathrm{BN}\!-\!\mathrm{ReLU}$}
layers; the first two blocks are followed by $2\times 2$ max pooling.
The final block is followed by global aggregation via adaptive average
pooling, producing a 256-dimensional vector, which is then projected to
the discriminant space $\mathbb{R}^{d}$ with $d=C-1$:
\begin{multline*}
\mathrm{Conv}(C_{\mathrm{in}}\!\to\!64)
\to
\mathrm{Conv}(64\!\to\!64)
\to
\mathrm{Pool}
\to
\mathrm{Conv}(64\!\to\!128)\\
\to
\mathrm{Conv}(128\!\to\!128)
\to
\mathrm{Pool}
\to
\mathrm{Conv}(128\!\to\!256)\\
\to
\mathrm{Conv}(256\!\to\!256)
\to
\mathrm{AdaptiveAvgPool}
\to
\mathrm{Linear}(256\!\to\!d),
\end{multline*}
where $C_{\mathrm{in}}=1$ for Fashion-MNIST and $C_{\mathrm{in}}=3$ for
CIFAR-10/100.  This encoder architecture is used unchanged across all experiments.
For the softmax baseline, the final linear layer also outputs $d=C-1$
features, which are then fed into a linear classifier
$\mathbb{R}^{d}\!\to\!\mathbb{R}^{C}$.

\paragraph*{Heads and training}
For each dataset we compare models built on the same encoder:
\begin{enumerate}
\item A linear softmax layer $Wz+b\in\mathbb{R}^C$ trained with cross-entropy.
\item An LDA head trained with DNLL, with different covariance parameterizations:
\begin{itemize}
  \item {spherical:} $\Sigma=\sigma^2 I_d$,
  \item {diagonal:} $\Sigma=\mathrm{diag}(\sigma_1^2,\ldots,\sigma_d^2)$,
  \item {full:} $\Sigma\succ 0$, parameterized via a Cholesky factor.
\end{itemize}
\end{enumerate}
We initialize the LDA parameters so that the latent classes start out reasonably well separated.
Priors are set to $\pi_c = 1/C$.
Each class mean $\mu_c$ is drawn from $\mathcal{N}\left(0,6^2/(2d)\right)$, placing the means in an isotropic cloud whose expected pairwise distance is on the order of~6, ensuring that different classes begin in distinct regions of the latent space.
The shared variance is set to $\sigma^2 = 1$, providing a non-degenerate initial scale.

\paragraph*{Results}
Table~\ref{tab:image-results} reports test accuracies averaged over three random seeds.
Across all three datasets, Deep LDA trained with DNLL is competitive with the softmax baseline.
Moreover, the covariance ablation shows that increasing covariance flexibility from spherical to diagonal or full
does not yield a consistent accuracy gain: diagonal and full $\Sigma$ perform similarly to spherical $\Sigma$.
This suggests that the main benefit of DNLL stems from the likelihood--separation interplay rather than from
a highly expressive covariance model, and that the lightweight spherical head is a robust default. 
We use it in the remaining experiments.

\begin{table}[htbp]
  \centering
  \caption{Image classification: test accuracies (mean $\pm$ 2\,std over 3 runs).}
  \label{tab:image-results}
  \begin{tabular}{llccc}
    \toprule
    Head & Loss & Fashion-MNIST & CIFAR-10 & CIFAR-100 \\
    \midrule
    Softmax & Cross Entropy 
    & $93.61\pm0.76$ 
    & $89.07\pm0.40$ 
    & $64.67\pm0.60$ \\
    Deep LDA (spherical $\Sigma$) & DNLL 
    & $92.82\pm0.30$ 
    & $90.33\pm0.56$ 
    & $66.09\pm1.30$ \\
    \midrule
    Deep LDA (diagonal $\Sigma$) & DNLL 
    & $92.84\pm0.52$ 
    & $90.09\pm0.68$ 
    & $65.95\pm0.62$ \\
    Deep LDA (full $\Sigma$) & DNLL 
    & $93.00\pm0.46$ 
    & $90.24\pm0.48$ 
    & $65.51\pm0.42$ \\  
    \bottomrule
  \end{tabular}
\end{table}




\begin{figure}[htbp]
  \centering
  \includegraphics[width=\textwidth]{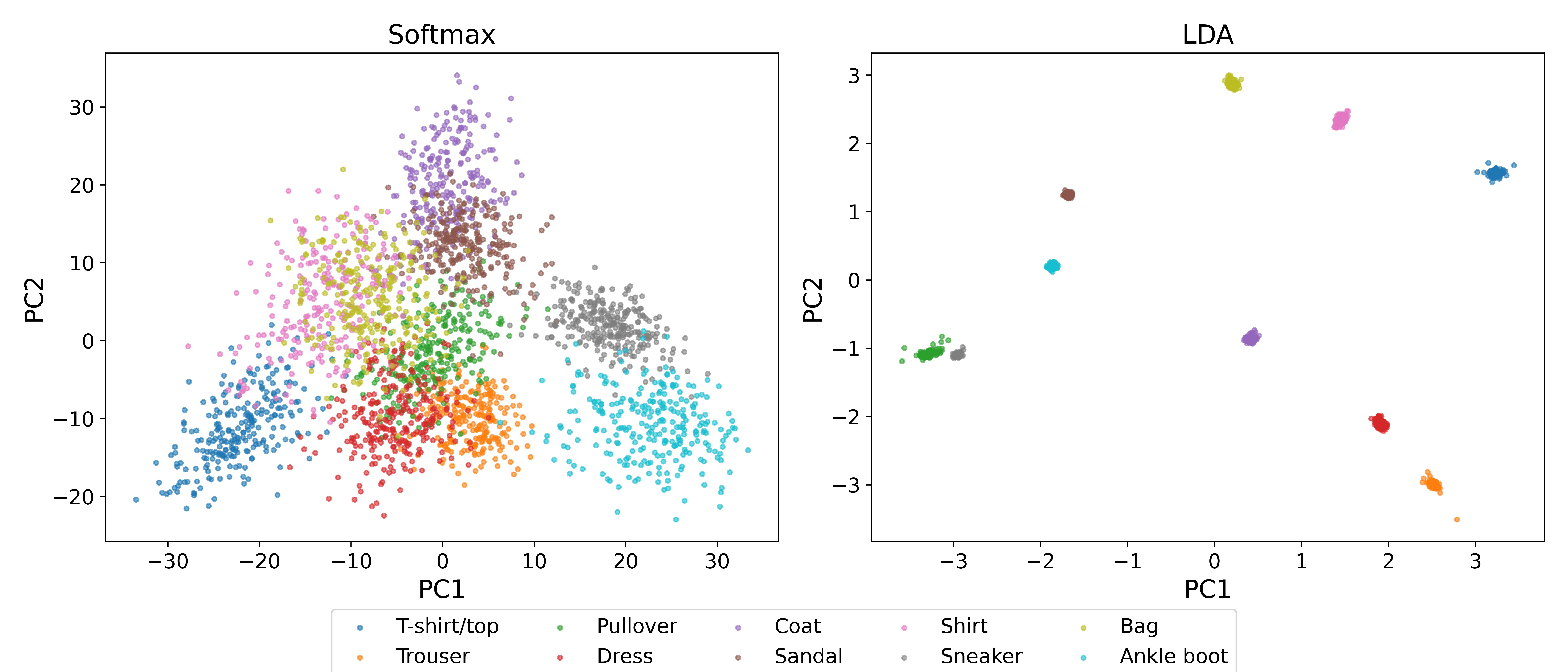}
  \includegraphics[width=\textwidth]{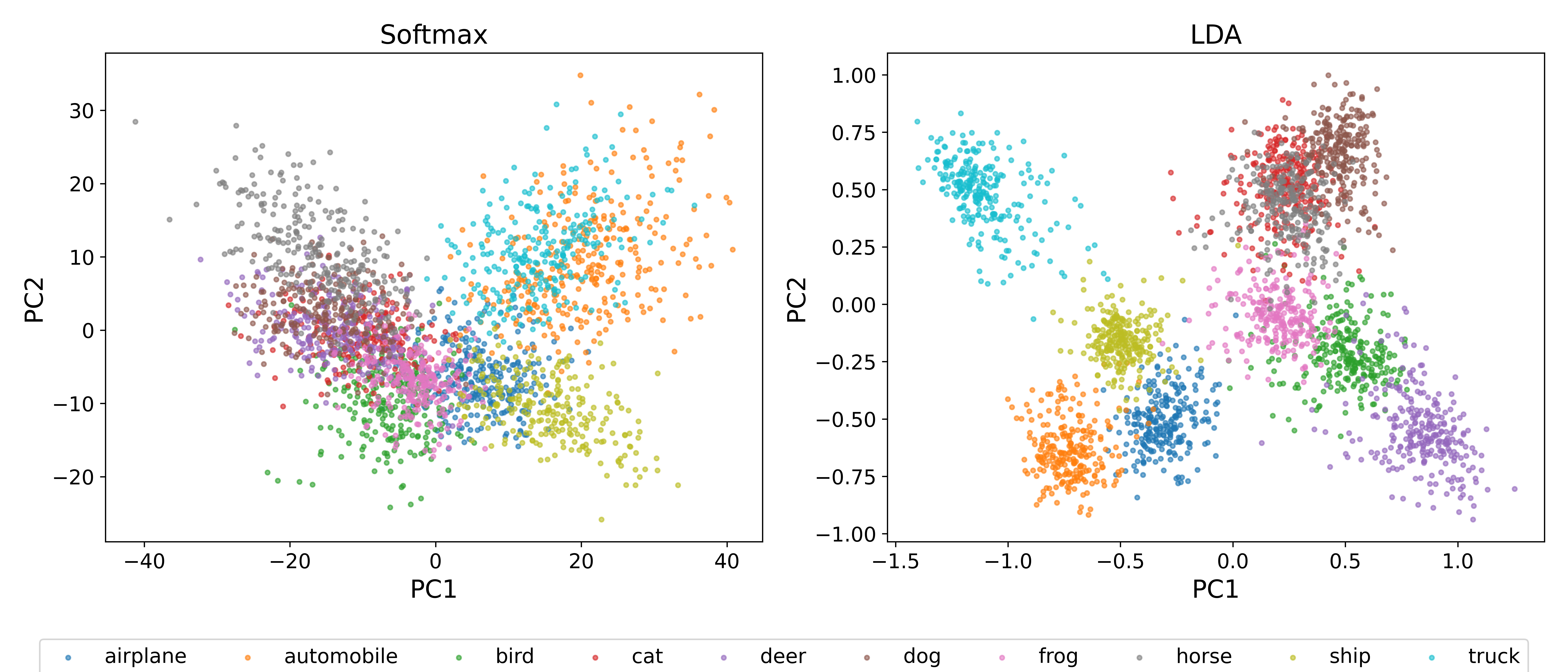}
  \caption{PCA projections of Fashion-MNIST (top) and CIFAR-10 (bottom) embeddings learned with the two classification heads. Deep LDA produces tighter, better-separated clusters even in the 2D projection.}
  \label{fig:img-embeddings-2d}
\end{figure}

\paragraph*{Embedding geometry}
To probe the geometry induced by Deep LDA, we visualize the learned
embeddings on Fahion-MNIST and CIFAR-10.  After training, we collect encoder outputs
$z_i=f_\psi(x_i)\in\mathbb{R}^{C-1}$ for a random subset of training
samples and project them to two dimensions using PCA.  The resulting
scatter plots are shown in Figure~\ref{fig:img-embeddings-2d}. 
Under
the softmax head, the projected clusters are clearly class-dependent
but partially overlapping.  In contrast, the Deep
LDA model trained with DNLL produces tight
clusters that remain better separated even in the 2D projection.  

These findings are consistent with the synthetic experiments of
Section~\ref{sec:mle} and \ref{sec:dnll}: DNLL encourages embeddings to
concentrate in regions where their own class density is high and the
mixture density is not too high, leading to clean, interpretable latent representations
without sacrificing classification accuracy.

\begin{figure}[htbp]
  \centering
  \includegraphics[width=0.85\linewidth]{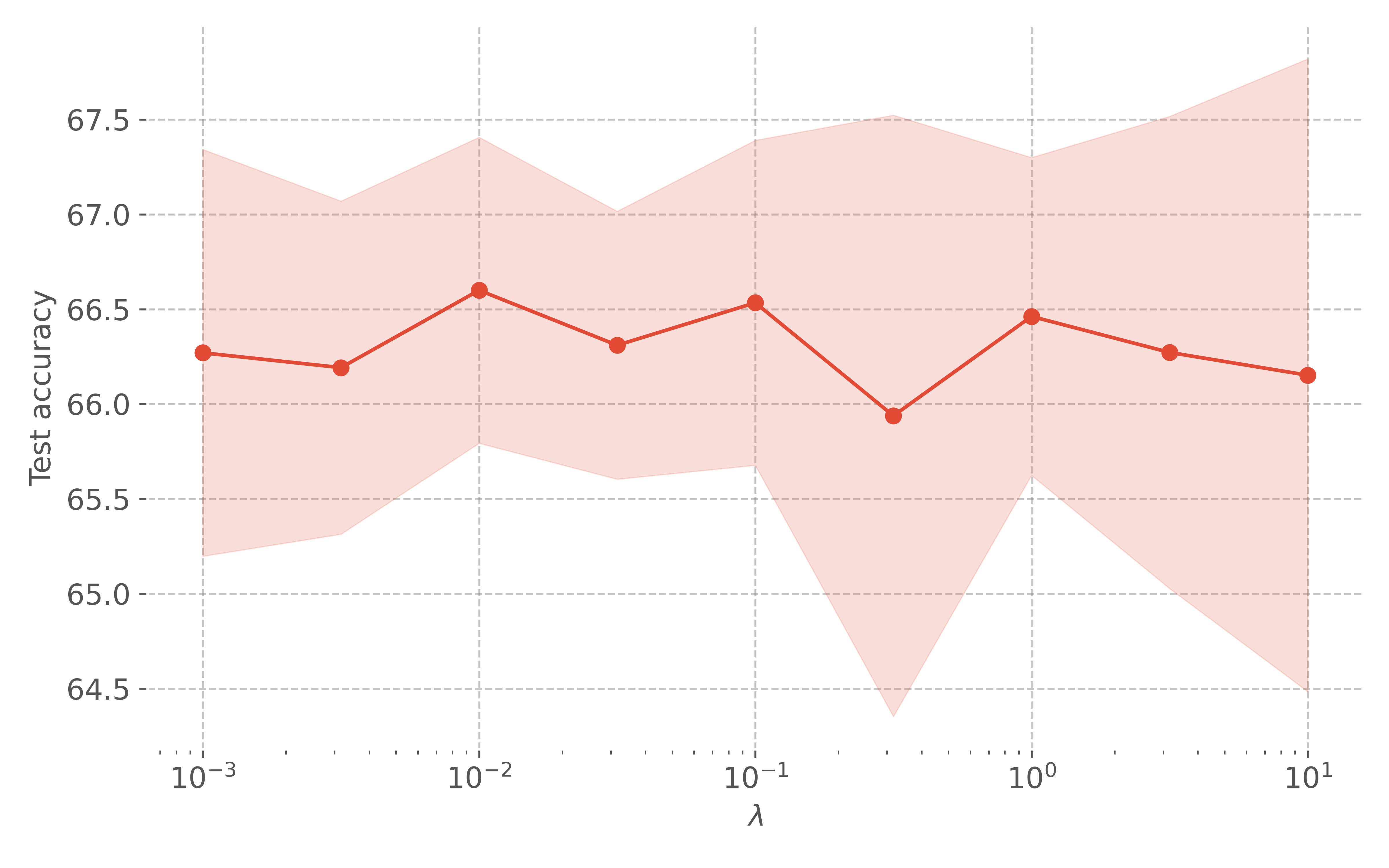}
  \caption{Sensitivity of Deep LDA  test accuracy to the regularization weight $\lambda$ on CIFAR-100 (mean $\pm$ 2 standard deviations  over 5 runs per each value of $\lambda$). Accuracy is stable across several orders of magnitude, supporting our use of a fixed $\lambda=0.01$ throughout.}
  \label{fig:lambda_acc}
\end{figure}

\paragraph{Sensitivity to the DNLL weight $\lambda$ (accuracy).}
All Deep LDA experiments in this paper use a fixed DNLL weight $\lambda=0.01$.
To assess whether the method is sensitive to this choice, we perform a sweep over $\lambda$ across several orders of magnitude on CIFAR-100 and report the 95\% confidence intervals for the mean test accuracy over 5 random seeds per each value of $\lambda$ (Figure~\ref{fig:lambda_acc}).
Test accuracy varies only modestly across the sweep, indicating that DNLL does not require delicate tuning of $\lambda$ for competitive classification performance.
We therefore use $\lambda=0.01$ in all experiments for simplicity and comparability.

Additional $\lambda$ sweeps on CIFAR-10 and Fashion-MNIST are reported in Appendix~\ref{app:lambda} and show the same qualitative behavior.

\subsection{Calibration and Confidence Behavior}\label{sec:calibration}

\begin{figure}[htbp]
    \centering

    \begin{subfigure}[t]{0.48\textwidth}
        \centering
        \includegraphics[width=\linewidth]{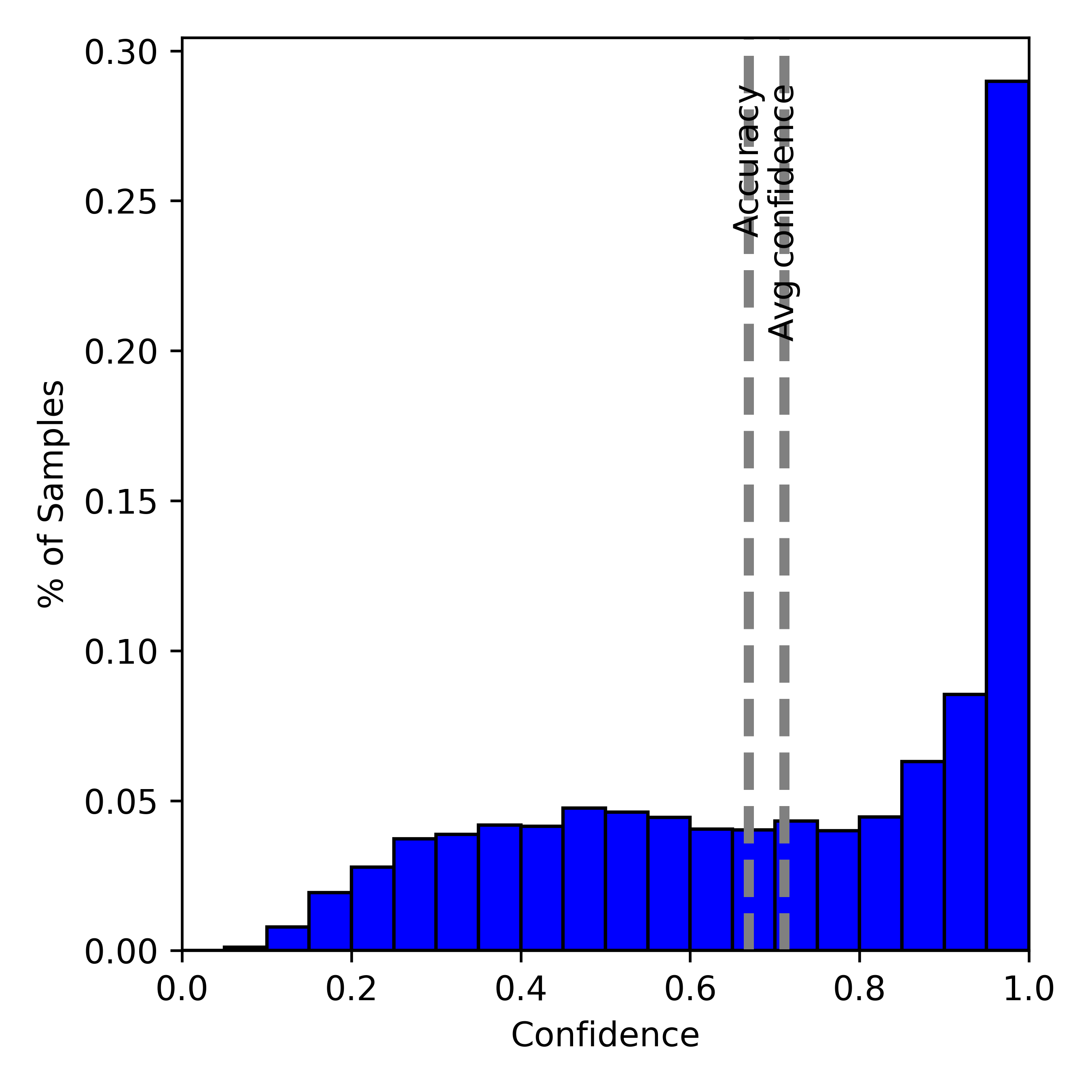}
        \caption{LDA head}
        \label{fig:cifar100_conf_hist_lda}
    \end{subfigure}
    \hfill
    \begin{subfigure}[t]{0.48\textwidth}
        \centering
        \includegraphics[width=\linewidth]{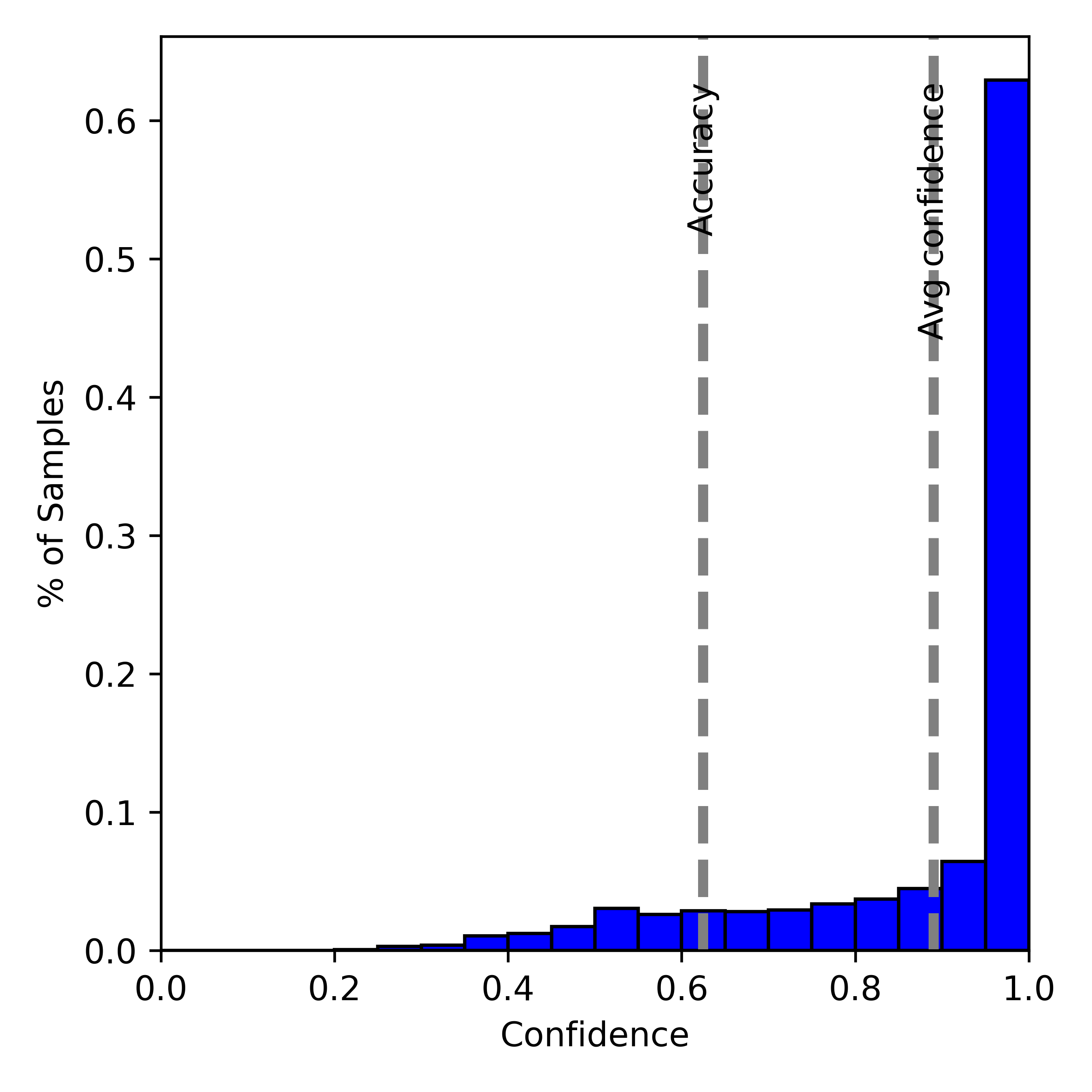}
        \caption{Softmax head}
        \label{fig:cifar100_conf_hist_softmax}
    \end{subfigure}

    \vspace{0.5em}

    \begin{subfigure}[t]{0.48\textwidth}
        \centering
        \includegraphics[width=\linewidth]{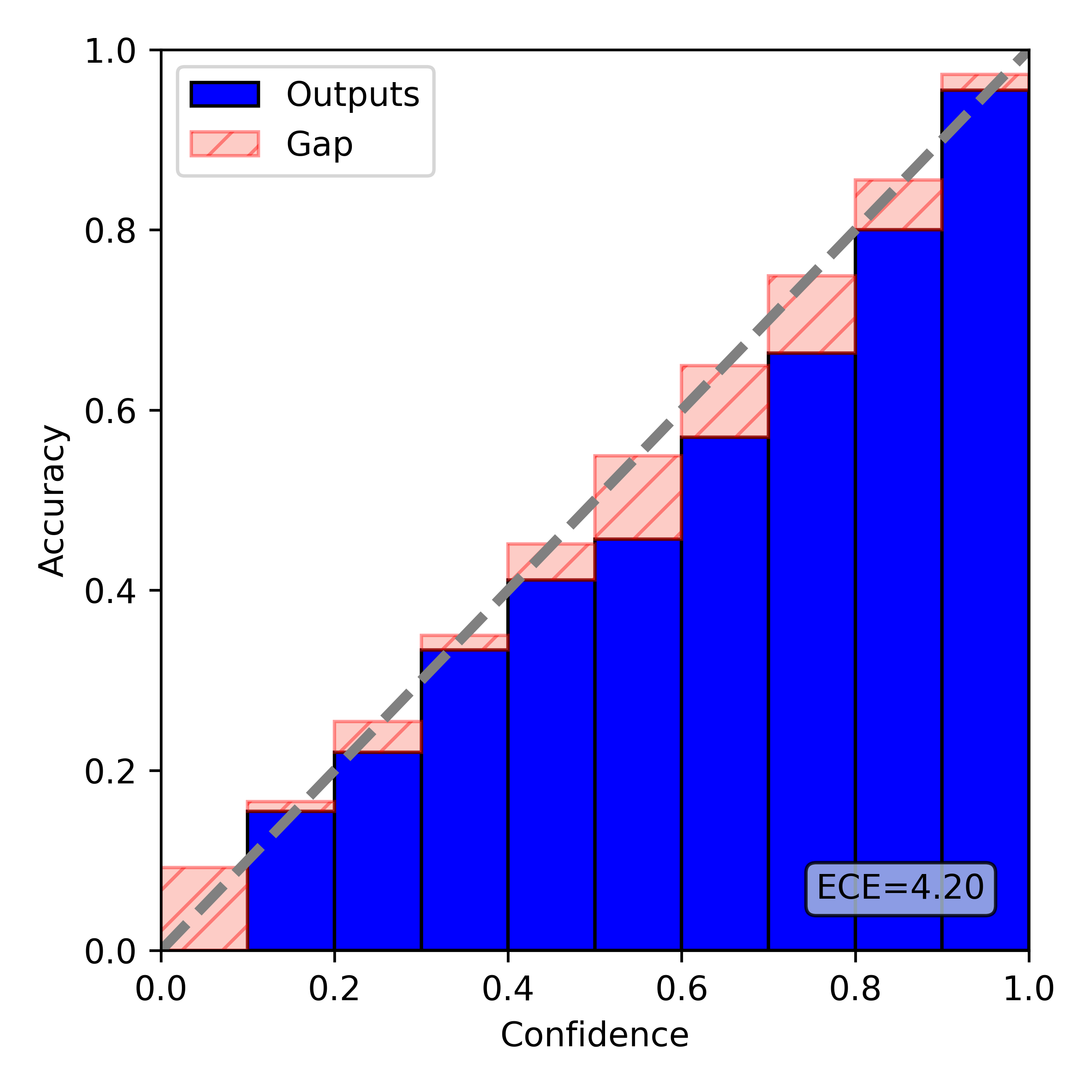}
        \caption{LDA head}
        \label{fig:cifar100_reliability_lda}
    \end{subfigure}
    \hfill
    \begin{subfigure}[t]{0.48\textwidth}
        \centering
        \includegraphics[width=\linewidth]{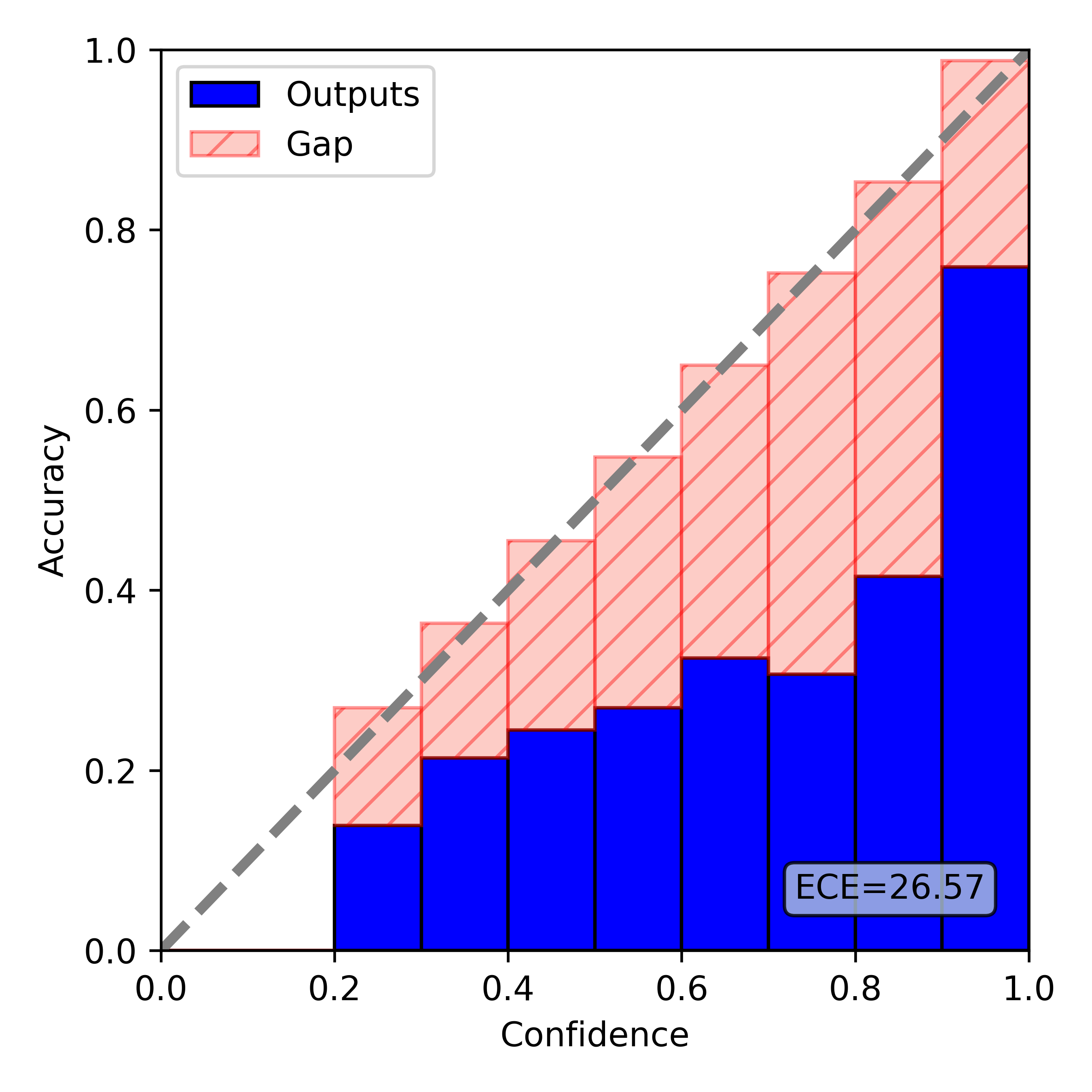}
        \caption{Softmax head}
        \label{fig:cifar100_reliability_softmax}
    \end{subfigure}

    \caption{
    \textbf{Calibration and confidence behavior on CIFAR-100.}
    Top row: histograms of maximum predicted confidence, with dashed vertical lines indicating average confidence and test accuracy.
    Bottom row: reliability diagrams with shaded calibration gaps and reported Expected Calibration Error (ECE).
    While the softmax classifier exhibits severe overconfidence and poor calibration (ECE $=26.6\%$),
    the LDA-based generative head trained with DNLL produces substantially better calibrated predictions (ECE $=4.2\%$),
    despite comparable classification accuracy.
    }
    \label{fig:cifar100_calibration}
\end{figure}

Beyond classification accuracy, we evaluate the \emph{calibration} of predictive probabilities, i.e., whether predicted confidences match empirical correctness frequencies.
For a test sample $x$, let $\hat{y}(x)=\arg\max_c p_\theta(c\mid x)$ and let the corresponding \emph{confidence} be
$\hat{p}(x)=\max_c p_\theta(c\mid x)$.
A \emph{confidence histogram} reports the distribution of $\hat{p}(x)$ over the test set; it reveals whether a model tends to make highly confident predictions (e.g., $\hat{p}(x)\approx 1$) or produces more moderate probabilities.

To quantify calibration, we use \emph{reliability diagrams} and the \emph{Expected Calibration Error (ECE)} of \cite{guo2017calibration}.
A reliability diagram bins test points by confidence, e.g., $B_m=\{x:\hat{p}(x)\in I_m\}$ for confidence intervals $I_m$, and plots, for each bin, the empirical accuracy
$\mathrm{acc}(B_m)=\frac{1}{|B_m|}\sum_{x\in B_m}\mathbb{I}\{\hat{y}(x)=y\}$
against the average confidence
$\mathrm{conf}(B_m)=\frac{1}{|B_m|}\sum_{x\in B_m}\hat{p}(x)$.
Perfect calibration corresponds to $\mathrm{acc}(B_m)\approx \mathrm{conf}(B_m)$ across bins (points lying on the diagonal).
ECE summarizes the overall deviation as
\begin{equation}
\mathrm{ECE} \;=\; \sum_{m=1}^M \frac{|B_m|}{n}\,\big|\mathrm{acc}(B_m)-\mathrm{conf}(B_m)\big|,
\end{equation}
where $n$ is the number of test samples and $M$ is the number of bins (we use $M=10$ with uniform-width bins).

Figure~\ref{fig:cifar100_calibration} compares an LDA-based generative head trained with DNLL against a standard softmax classifier on CIFAR-100.
While both models achieve similar classification accuracy, their confidence behavior differs markedly.
The softmax head exhibits severe overconfidence, assigning near-unit confidence to a large fraction of samples despite substantially lower empirical accuracy, resulting in an ECE of 26.6\%.
In contrast, the LDA head produces a broader confidence distribution and substantially improved alignment between confidence and accuracy, achieving an ECE of 4.2\%.

Appendix C reports the same confidence histograms and reliability diagrams on CIFAR-10 and Fashion-MNIST. The LDA head remains consistently better calibrated than softmax (ECE 3.08\% vs 7.73\% on CIFAR-10; 2.98\% vs 5.05\% on Fashion-MNIST).

These results indicate that preserving an explicit generative structure in the classifier head leads to significantly better calibrated predictive probabilities, even in standard supervised classification settings.

\begin{figure}[htbp]
  \centering
  \includegraphics[width=0.85\linewidth]{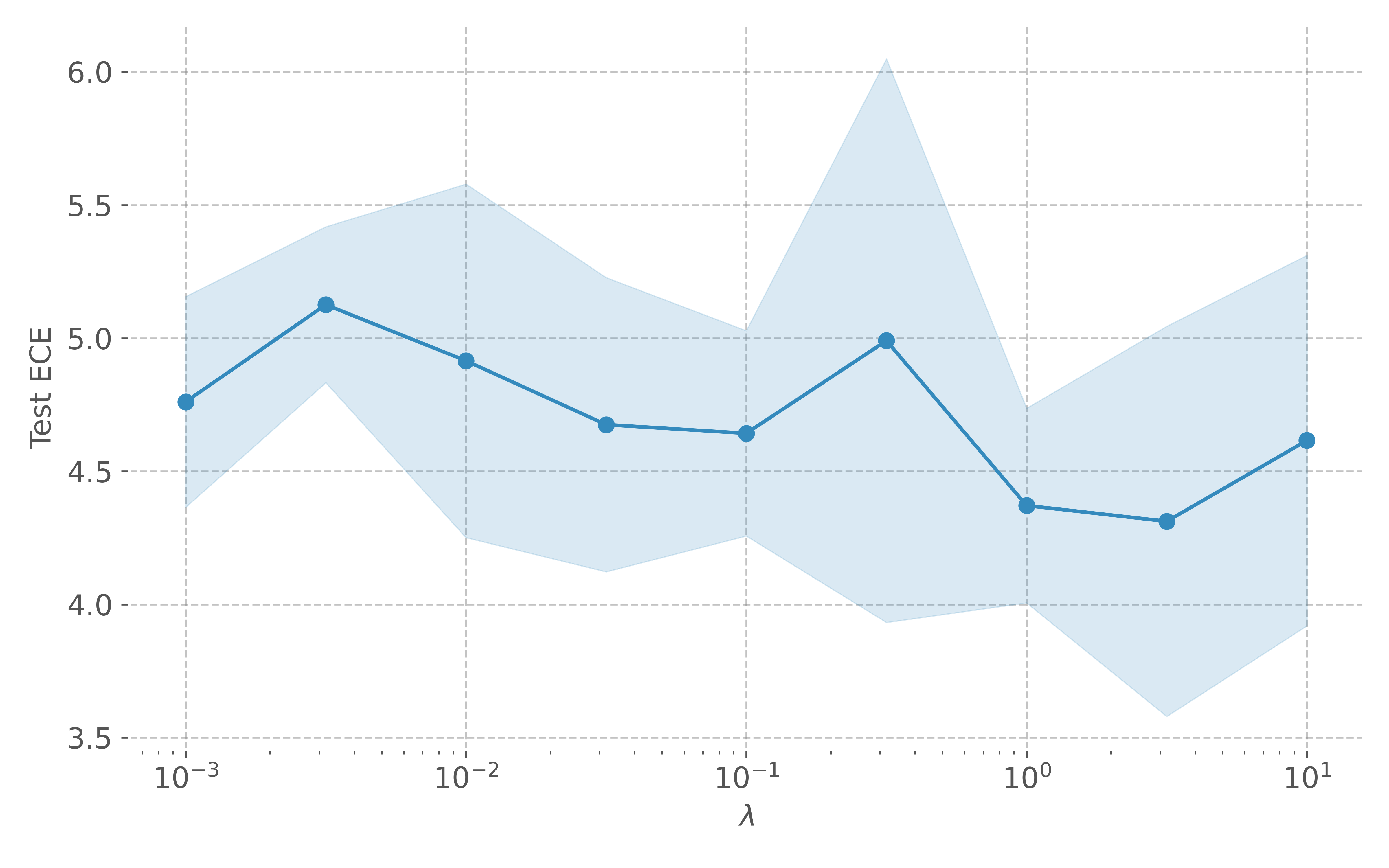}
  \caption{Sensitivity of calibration to the DNLL weight $\lambda$ on CIFAR-100, measured by Expected Calibration Error (ECE) (means $\pm$ 2 standard deviations over 5 runs per each value of $\lambda$). Calibration remains stable across several orders of magnitude.}
  \label{fig:lambda_ece}
\end{figure}

\paragraph{Sensitivity to $\lambda$ (calibration).}
Using the same CIFAR-100 sweep as in Figure~\ref{fig:lambda_acc},
we also examine how calibration depends on $\lambda$.
Figure~\ref{fig:lambda_ece} reports ECE as a function of $\lambda$; calibration remains consistently strong across a wide range of values.
This indicates that the calibration benefits of DNLL are not tied to a finely tuned $\lambda$.

\subsection{What does the DNLL weight $\lambda$ control?}
\label{sec:lambda-controls}

Figures~\ref{fig:lambda_acc} and~\ref{fig:lambda_ece} show that, on CIFAR-100, both test accuracy and calibration (ECE) are remarkably
insensitive to the DNLL weight $\lambda$ across several orders of magnitude. 
This naturally raises the question: \emph{if $\lambda$ does not change accuracy or ECE, what does it
actually affect?}

\paragraph{A gradient-based interpretation.}
Recall the definition of the DNLL loss \eqref{eq:dnll}.
Let $\delta(z)=(\delta_1(z),\ldots,\delta_C(z))^\top$ and define $\exp(\delta(z))$ elementwise.
A direct derivative computation gives the gradient with respect to the vector of discriminants:
\begin{equation}
\nabla_{\delta}\,\ell_{\mathrm{DNLL}}(z,y)
\;=\;
\lambda\,\exp\!\big(\delta(z)\big)
\;-\;
e_y,
\label{eq:dnll-grad-delta}
\end{equation}
where $e_y$ is the $y$-th standard basis vector in $\mathbb{R}^C$.
Equivalently, componentwise,
\begin{equation}
\frac{\partial \ell_{\mathrm{DNLL}}(z,y)}{\partial \delta_c(z)}
=
\begin{cases}
-1 + \lambda \exp(\delta_y(z)), & c=y,\\[2pt]
\lambda \exp(\delta_c(z)), & c\neq y.
\end{cases}
\end{equation}
Thus, gradient-based optimization pushes $\lambda \exp(\delta_y(z))$ toward $1$ (for the true class)
and $\lambda \exp(\delta_c(z))$ toward $0$ (for competing classes). In particular, for typical
training embeddings one expects $\exp(\delta_y(z))$ to scale on the order of $1/\lambda$.

\paragraph{Why this targets the covariance scale in the spherical head.}
For the LDA head,
\begin{equation}
\exp\!\big(\delta_c(z)\big)
=
\pi_c\,\varphi(z;\mu_c,\Sigma)\cdot(2\pi)^{d/2},
\end{equation}
so $\exp(\delta_c(z))$ is proportional to the class-conditional Gaussian density evaluated at $z$.
In our main image-classification configuration we use a spherical shared covariance
$\Sigma=\sigma^2 I_d$.
Under this parameterization, the \emph{only} scalar degree of freedom that globally controls the
``peakiness'' (overall scale) of $\varphi(z;\mu_c,\sigma^2 I_d)$ is $\sigma^2$.
Consequently, the stationarity condition suggested by~\eqref{eq:dnll-grad-delta},
$\exp(\delta_y(z))\approx 1/\lambda$, predicts that decreasing $\lambda$ should be accompanied by
a decrease in the learned $\sigma$ (sharper class-conditionals), while increasing $\lambda$ should
allow larger $\sigma$ (broader class-conditionals).

\begin{figure}[t]
  \centering
  \includegraphics[width=0.72\linewidth]{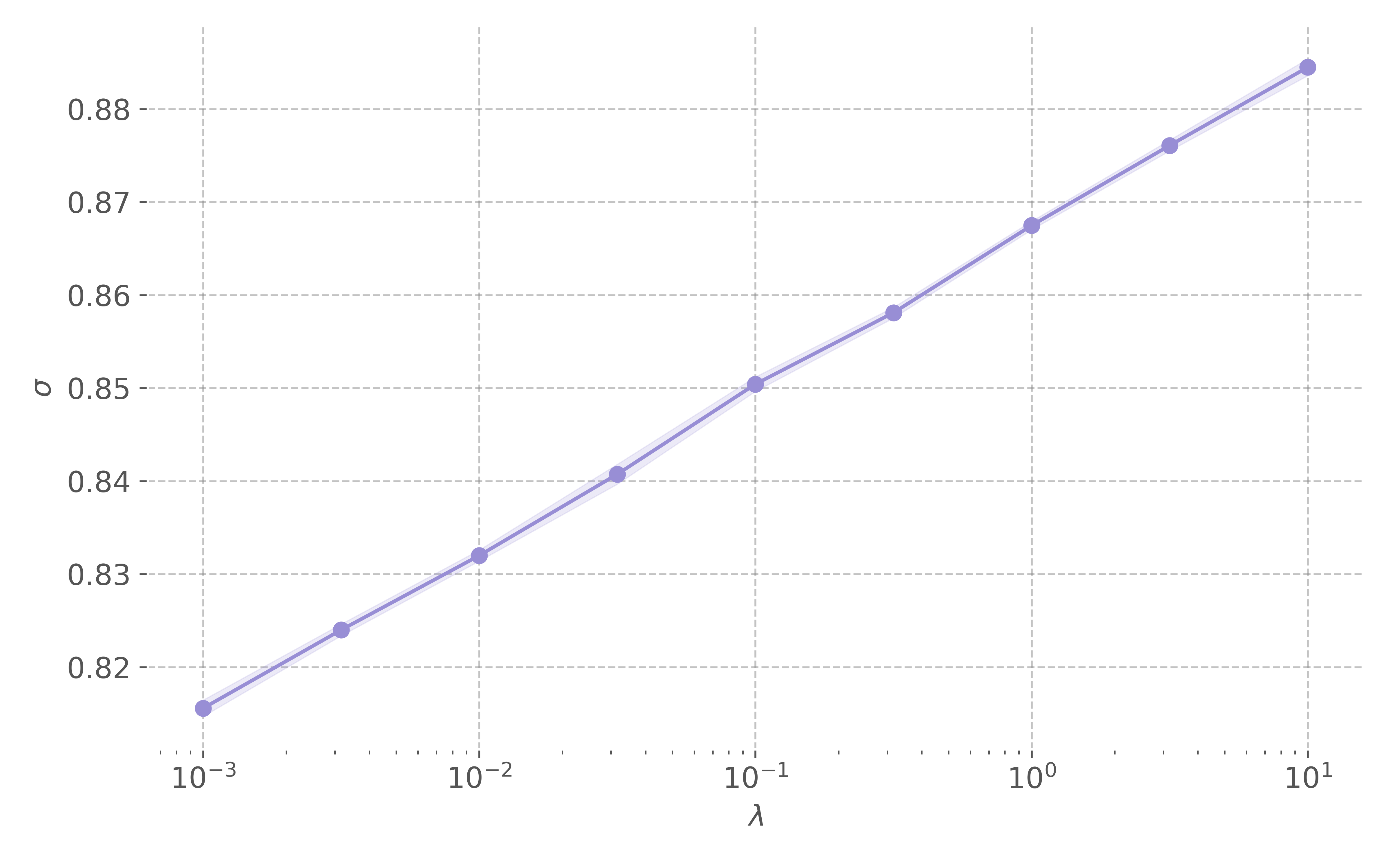}
  \caption{Dependence of the learned spherical covariance scale $\sigma$ to the DNLL weight $\lambda$
  on CIFAR-100 (same sweep protocol as Figures~6 and~8). While accuracy and ECE are nearly invariant
  to $\lambda$, the learned $\sigma$ changes systematically: decreasing $\lambda$ results in smaller
  $\sigma$ (sharper class-conditionals).}
  \label{fig:lambda-vs-sigma}
\end{figure}

\paragraph{Empirical confirmation.}
To test this hypothesis, we reran the same CIFAR-100 $\lambda$ sweep used in
Figures~\ref{fig:lambda_acc} and~\ref{fig:lambda_ece} and recorded the final learned value of $\sigma$ (for the spherical head) at the end
of training.
Figure~\ref{fig:lambda-vs-sigma} reports $\sigma$ as a function of $\lambda$ (mean over runs).
In contrast to accuracy and ECE, the covariance scale exhibits a clear monotone dependence on
$\lambda$: smaller $\lambda$ consistently produces smaller $\sigma$ (sharper Gaussians), whereas
larger $\lambda$ yields larger $\sigma$ (broader Gaussians).
This explains why $\lambda$ can vary widely without affecting classification metrics: in this regime,
$\lambda$ primarily acts as a \emph{scale/temperature knob} for the generative head (through $\sigma$),
rather than as a parameter that strongly changes the induced decision rule.

The same monotone $\sigma$-$\lambda$ relationship appears on CIFAR-10 and Fashion-MNIST (Appendix~\ref{app:lambda}).

\section{Conclusion}

We revisited Deep LDA through the lens of likelihood-based training and showed that
unconstrained maximum-likelihood optimization admits pathological solutions in which
class clusters overlap, while purely discriminative cross-entropy training breaks the
probabilistic interpretation of the LDA head.
To reconcile these two perspectives, we introduced the Discriminative Negative
Log-Likelihood (DNLL), which augments the LDA log-likelihood with a density penalty
that explicitly discourages regions where several classes have simultaneously high
probability.
On synthetic data and standard image benchmarks, Deep LDA with DNLL produces clean,
well-separated latent class clusters and achieves test accuracy competitive with
softmax classifiers, while yielding substantially better calibrated predictive
probabilities.

Our sensitivity analysis further shows that DNLL does not require delicate tuning of
the regularization weight $\lambda$: both accuracy and calibration (ECE) remain stable
across several orders of magnitude.
At the same time, $\lambda$ is not inert: in the spherical-head setting
$\Sigma=\sigma^2 I$, varying $\lambda$ systematically changes the learned covariance
scale $\sigma^2$ (smaller $\lambda$ leads to smaller $\sigma$, i.e., sharper Gaussian
class-conditionals), even when downstream metrics remain essentially unchanged.
This suggests that, in practice, $\lambda$ primarily acts as a scale/temperature knob
for the generative head---controlling the concentration of the fitted class densities---rather than as a parameter that strongly affects the induced classifier.

These findings reinforce the main message of the paper: simple generative structure,
combined with an explicitly discriminative loss, can restore the viability of
likelihood-based training for deep discriminant models while producing meaningful
predictive probabilities.
An interesting direction for future work is to make this ``scale'' effect explicit,
e.g., by reparameterizing DNLL to absorb $\lambda$ into a temperature (or variance)
parameter, or by designing adaptive schemes that target a desired covariance scale or
confidence profile during training.

\paragraph{Reproducibility} Code to reproduce our experiments is available at \url{https://github.com/zh3nis/DNLL}.

\paragraph{AI Assistance} We used an AI-based tool for editorial polishing of the text; the authors take full responsibility for the content.

\bibliography{ref}

@inproceedings{guo2017calibration,
  title={On calibration of modern neural networks},
  author={Guo, Chuan and Pleiss, Geoff and Sun, Yu and Weinberger, Kilian Q},
  booktitle={International conference on machine learning},
  pages={1321--1330},
  year={2017},
  organization={PMLR}
}

@article{fisher1936use,
  title={The use of multiple measurements in taxonomic problems},
  author={Fisher, Ronald A},
  journal={Annals of eugenics},
  volume={7},
  number={2},
  pages={179--188},
  year={1936},
  publisher={Wiley Online Library}
}

@article{rao1948,
 ISSN = {00359246},
 URL = {http://www.jstor.org/stable/2983775},
 author = {C. Radhakrishna Rao},
 journal = {Journal of the Royal Statistical Society. Series B (Methodological)},
 number = {2},
 pages = {159--203},
 publisher = {[Royal Statistical Society, Wiley]},
 title = {The Utilization of Multiple Measurements in Problems of Biological Classification},
 urldate = {2024-01-30},
 volume = {10},
 year = {1948}
}

@inproceedings{mika1999fisher,
  title={Fisher discriminant analysis with kernels},
  author={Mika, Sebastian and Ratsch, Gunnar and Weston, Jason and Scholkopf, Bernhard and Mullers, Klaus-Robert},
  booktitle={Neural networks for signal processing IX: Proceedings of the 1999 IEEE signal processing society workshop (cat. no. 98th8468)},
  pages={41--48},
  year={1999},
  organization={Ieee}
}

@article{hastie1996discriminant,
  title={Discriminant analysis by Gaussian mixtures},
  author={Hastie, Trevor and Tibshirani, Robert},
  journal={Journal of the Royal Statistical Society Series B: Statistical Methodology},
  volume={58},
  number={1},
  pages={155--176},
  year={1996},
  publisher={Oxford University Press}
}

@inproceedings{DBLP:journals/corr/DorferKW15,
  author       = {Matthias Dorfer and
                  Rainer Kelz and
                  Gerhard Widmer},
  editor       = {Yoshua Bengio and
                  Yann LeCun},
  title        = {Deep Linear Discriminant Analysis},
  booktitle    = {4th International Conference on Learning Representations, {ICLR} 2016,
                  San Juan, Puerto Rico, May 2-4, 2016, Conference Track Proceedings},
  year         = {2016},
  url          = {http://arxiv.org/abs/1511.04707},
  bibsource    = {dblp computer science bibliography, https://dblp.org}
}

@inproceedings{DBLP:journals/corr/KingmaB14,
  author       = {Diederik P. Kingma and
                  Jimmy Ba},
  editor       = {Yoshua Bengio and
                  Yann LeCun},
  title        = {Adam: {A} Method for Stochastic Optimization},
  booktitle    = {3rd International Conference on Learning Representations, {ICLR} 2015,
                  San Diego, CA, USA, May 7-9, 2015, Conference Track Proceedings},
  year         = {2015},
  url          = {http://arxiv.org/abs/1412.6980},
  bibsource    = {dblp computer science bibliography, https://dblp.org}
}

@article{xiao2017fashion,
  title={Fashion-mnist: a novel image dataset for benchmarking machine learning algorithms},
  author={Xiao, Han and Rasul, Kashif and Vollgraf, Roland},
  journal={arXiv preprint arXiv:1708.07747},
  year={2017}
}

@article{krizhevsky2009learning,
  title={Learning multiple layers of features from tiny images},
  author={Krizhevsky, Alex and Hinton, Geoffrey and others},
  year={2009},
  publisher={Toronto, ON, Canada}
}

@article{ACU2021125578,
title = {Information potential for some probability density functions},
journal = {Applied Mathematics and Computation},
volume = {389},
pages = {125578},
year = {2021},
issn = {0096-3003},
author = {Ana-Maria Acu and Gülen Başcanbaz-Tunca and Ioan Rasa},
}

\clearpage
\appendix

\section{Analysis of the DNLL density penalty for the LDA head}
\label{app:dnll-detailed}

This appendix provides the detailed information potential analysis of
 the LDA head used in the paper. We (i) derive a closed-form expression for the information potential
of the LDA marginal $p_\theta(z)$, (ii) interpret the resulting term as a
repulsive interaction between class means and a barrier against covariance
collapse, and (iii) relate the empirical DNLL density penalty
$\mathbb E_{Z\sim p}[p_\theta(Z)]$ to the intrinsic information potential
$\int p_\theta(z)^2dz$ via a KL-controlled bound. We also record a spherical
specialization $\Sigma=\sigma^2 I_d$.

\subsection{Notation and the LDA mixture in embedding space}
\label{app:notation}

In the main text, the encoder $f_\psi$ maps inputs to embeddings
\[
z=f_\psi(x)\in\mathbb R^d.
\]
The LDA head is a shared-covariance Gaussian class-conditional model
\[
p_\theta(z\mid y=c)=\mathcal N(z;\mu_c,\Sigma),
\qquad
p_\theta(y=c)=\pi_c,
\qquad
\Sigma\succ0,\quad \pi_c>0,\quad \sum_{c=1}^C\pi_c=1,
\]
where $\theta=(\pi,\{\mu_c\}_{c=1}^C,\Sigma)$.
The discriminant score used in the paper is
\[
\delta_c(z)=\log p_\theta(z,y=c)=\log\pi_c+\log\mathcal N(z;\mu_c,\Sigma).
\]
The induced marginal (mixture) density in embedding space is
\begin{equation}
\label{eq:app-mixture}
p_\theta(z)=\sum_{c=1}^C \pi_c\,\varphi_c(z),
\qquad
\varphi_c(z)=\mathcal N(z;\mu_c,\Sigma).
\end{equation}
Note that $\sum_{c=1}^C e^{\delta_c(z)}=\sum_c p_\theta(z,c)=p_\theta(z)$, so the
DNLL penalty term in the main text is (up to constants) the mixture density
evaluated at the embedding.

\subsection{Information potential of the LDA mixture}
\label{app:collision-prob}

Define the information potential
\begin{equation}
\label{eq:app-collision-def}
C(p_\theta)\;:=\;\int_{\mathbb R^d} p_\theta(z)^2\,dz.
\end{equation}

\begin{theorem}[Information potential for the LDA mixture (shared covariance)]
\label{thm:app-collision-lda}
Let $p_\theta(z)$ be the LDA mixture \eqref{eq:app-mixture}. Then
\begin{equation}
\label{eq:app-collision-lda}
C(p_\theta)
=
\sum_{i,j=1}^C \pi_i\pi_j\,\mathcal N(\mu_i-\mu_j;0,2\Sigma).
\end{equation}
Equivalently,
\begin{equation}
\label{eq:app-collision-lda-explicit}
C(p_\theta)
=
(4\pi)^{-d/2}\,(\det\Sigma)^{-1/2}
\sum_{i,j=1}^C \pi_i\pi_j\,
\exp\!\Big(
-\tfrac14(\mu_i-\mu_j)^\top\Sigma^{-1}(\mu_i-\mu_j)
\Big).
\end{equation}
\end{theorem}

\begin{remark}[Diagonal vs.\ off-diagonal terms: repulsion and collapse]
\label{rem:app-repulsion}
The diagonal terms in \eqref{eq:app-collision-lda} are
\[
\pi_i^2\,\mathcal N(0;0,2\Sigma)
=(4\pi)^{-d/2}(\det\Sigma)^{-1/2}\,\pi_i^2,
\]
which do not depend on $\mu_i$ but \emph{diverge} as $\det\Sigma\downarrow 0$.
Thus, penalizing $C(p_\theta)$ discourages covariance collapse.

For $i\neq j$, the $(i,j)$ term equals
\[
\pi_i\pi_j\,(4\pi)^{-d/2}(\det\Sigma)^{-1/2}\,
\exp\!\Big(-\tfrac14\|\mu_i-\mu_j\|_{\Sigma^{-1}}^2\Big),
\]
which is \emph{repellant}: it decays exponentially in the Mahalanobis distance
between class centers. Hence, penalizing information potential can be interpreted
as introducing a repulsive potential between class means (while also penalizing
small determinant of $\Sigma$).
\end{remark}

\begin{proof}[Proof of Theorem~\ref{thm:app-collision-lda}]
By linearity,
\begin{align*}
\int p_\theta(z)^2\,dz
&=
\int\Big(\sum_{i=1}^C\pi_i\varphi_i(z)\Big)
     \Big(\sum_{j=1}^C\pi_j\varphi_j(z)\Big)\,dz
=
\sum_{i,j=1}^C \pi_i\pi_j\int \varphi_i(z)\varphi_j(z)\,dz.
\end{align*}
Thus we need to compute
\[
I_{ij}:=\int_{\mathbb R^d}\varphi_i(z)\varphi_j(z)\,dz,
\qquad
\varphi_i(z)=\mathcal N(z;\mu_i,\Sigma),\ \varphi_j(z)=\mathcal N(z;\mu_j,\Sigma).
\]

\paragraph{Step 1: reduction to convolution at zero.}
Define the reflected density
\[
\tilde\varphi_j(z):=\varphi_j(-z).
\]
Then, by a change of variables,
\begin{align*}
(\varphi_i * \tilde\varphi_j)(0)
&=
\int_{\mathbb R^d}\varphi_i(t)\tilde\varphi_j(-t)\,dt
=
\int_{\mathbb R^d}\varphi_i(t)\varphi_j(t)\,dt
=
I_{ij}.
\end{align*}
Hence,
\begin{equation}
\label{eq:app-Iij-conv}
I_{ij}=(\varphi_i * \tilde\varphi_j)(0).
\end{equation}

\paragraph{Step 2: reflection of a Gaussian.}
If $\varphi_j(z)=\mathcal N(z;\mu_j,\Sigma)$, then
\[
\tilde\varphi_j(z)=\varphi_j(-z)=\mathcal N(z;-\mu_j,\Sigma),
\]
since replacing $z$ by $-z$ flips the mean sign but leaves the covariance
unchanged.

\paragraph{Step 3: convolution of Gaussians.}
A standard identity states that for Gaussian densities,
\[
\mathcal N(\cdot;\mu_1,\Sigma_1) * \mathcal N(\cdot;\mu_2,\Sigma_2)
=
\mathcal N(\cdot;\mu_1+\mu_2,\Sigma_1+\Sigma_2).
\]
Applying it with $\varphi_i=\mathcal N(\cdot;\mu_i,\Sigma)$ and
$\tilde\varphi_j=\mathcal N(\cdot;-\mu_j,\Sigma)$ yields
\[
(\varphi_i * \tilde\varphi_j)(z)
=
\mathcal N(z;\mu_i-\mu_j,2\Sigma).
\]

\paragraph{Step 4: evaluation at $z=0$ and symmetry.}
Evaluating at $z=0$ gives
\[
I_{ij}=(\varphi_i * \tilde\varphi_j)(0)=\mathcal N(0;\mu_i-\mu_j,2\Sigma).
\]
By symmetry of the Gaussian density,
\[
\mathcal N(0;\mu_i-\mu_j,2\Sigma)=\mathcal N(\mu_i-\mu_j;0,2\Sigma).
\]
Substituting $I_{ij}$ into the expansion completes the derivation of
\eqref{eq:app-collision-lda}. Writing out the Gaussian density yields
\eqref{eq:app-collision-lda-explicit}.
\end{proof}

\subsection{Our losses: intrinsic vs.\ empirical overlap}
\label{app:our-losses}

In population form, consider the two objectives (written in embedding space):
\begin{align}
I_1(\theta)
&=
-\mathbb E_{(Z,Y)\sim p}\big[\log p_\theta(Z,Y)\big]
+
\lambda\int_{\mathbb R^d} p_\theta(z)^2\,dz,
\label{eq:app-I1}
\\
I_2(\theta)
&=
-\mathbb E_{(Z,Y)\sim p}\big[\log p_\theta(Z,Y)\big]
+
\lambda\int_{\mathbb R^d} p(z)\,p_\theta(z)\,dz,
\label{eq:app-I2}
\end{align}
where $p$ denotes the (unknown) data distribution of $(Z,Y)$ induced by the
encoder, and $p(z)$ its marginal in embedding space.
The DNLL penalty used in training is the empirical approximation of
$\int p(z)p_\theta(z)\,dz$:
with a training set $\{(z_n,y_n)\}_{n=1}^N$,
\[
\int p(z)p_\theta(z)\,dz
\approx
\frac1N\sum_{n=1}^N p_\theta(z_n)
=
\frac1N\sum_{n=1}^N\sum_{c=1}^C p_\theta(z_n,c)
=
\frac1N\sum_{n=1}^N\sum_{c=1}^C e^{\delta_c(z_n)}.
\]

\paragraph{Rewriting via KL.}
Assume $p(z)$ is a density and $p(y\mid z)$ a conditional pmf.
Up to an additive constant (the differential entropy of $p(z)$ plus the expected
Shannon entropy of $p(y\mid Z)$), the negative log-likelihood term can be written as
a KL divergence. Concretely,
\[
-\mathbb E_{(Z,Y)\sim p}\big[\log p_\theta(Z,Y)\big]
=
\mathrm{KL}\big(p(z,y)\,\|\,p_\theta(z,y)\big) + \text{const}.
\]
Therefore, up to an additive constant, \eqref{eq:app-I1}--\eqref{eq:app-I2} are equivalent to
\begin{align}
L_1(\theta)
&=
\mathrm{KL}\big(p(z,y)\,\|\,p_\theta(z,y)\big)
+
\lambda\int p_\theta(z)^2\,dz,
\label{eq:app-L1}
\\
L_2(\theta)
&=
\mathrm{KL}\big(p(z,y)\,\|\,p_\theta(z,y)\big)
+
\lambda\int p(z)\,p_\theta(z)\,dz.
\label{eq:app-L2}
\end{align}
We address: \emph{when does minimizing $L_2$ also minimize $L_1$?}

\subsection{A general overlap bound via KL}
\label{app:overlap-general}

\begin{proposition}[Control of overlap difference via KL]
\label{prop:app-overlap-kl}
Let $p$ and $p_\theta$ be probability densities on $\mathbb R^d$.
Assume $p_\theta\in L^\infty(\mathbb R^d)$ with $\|p_\theta\|_\infty<\infty$.
Then
\begin{equation}
\label{eq:app-overlap-kl}
\left|
\int_{\mathbb R^d} p(z)\,p_\theta(z)\,dz
-
\int_{\mathbb R^d} p_\theta(z)^2\,dz
\right|
\le
\|p_\theta\|_\infty\,\sqrt{2\,\mathrm{KL}(p(z)\|p_\theta(z))}.
\end{equation}
\end{proposition}

\begin{proof}
We begin by writing the difference as
\[
\int p(z)p_\theta(z)\,dz - \int p_\theta(z)^2\,dz
=
\int (p(z)-p_\theta(z))\,p_\theta(z)\,dz.
\]
By H\"older's inequality,
\[
\left|\int (p-p_\theta)p_\theta\right|
\le
\|p-p_\theta\|_1\,\|p_\theta\|_\infty
=
2\,\mathrm{TV}(p,p_\theta)\,\|p_\theta\|_\infty,
\]
where $\mathrm{TV}(p,p_\theta)=\tfrac12\|p-p_\theta\|_1$.
Applying Pinsker's inequality,
\[
\mathrm{TV}(p,p_\theta)\le \sqrt{\tfrac12\,\mathrm{KL}(p\|p_\theta)},
\]
we obtain \eqref{eq:app-overlap-kl}.
\end{proof}

\subsection{Specialization to the LDA mixture: an explicit constant}
\label{app:overlap-lda}

\begin{proposition}[Explicit overlap bound for the LDA head]
\label{prop:app-overlap-lda}
Let $p_\theta(z)=\sum_{c=1}^C \pi_c\mathcal N(z;\mu_c,\Sigma)$ with $\Sigma\succ0$.
Then
\begin{equation}
\label{eq:app-linfty-lda}
\|p_\theta\|_\infty \le (2\pi)^{-d/2}(\det\Sigma)^{-1/2},
\end{equation}
and hence
\begin{equation}
\label{eq:app-overlap-lda}
\left|
\int p(z)p_\theta(z)\,dz
-
\int p_\theta(z)^2\,dz
\right|
\le
(2\pi)^{-d/2}(\det\Sigma)^{-1/2}\,
\sqrt{2\,\mathrm{KL}(p(z)\|p_\theta(z))}.
\end{equation}
\end{proposition}

\begin{proof}
For each Gaussian component,
\[
\phi_c(z):=\mathcal N(z;\mu_c,\Sigma)
=(2\pi)^{-d/2}(\det\Sigma)^{-1/2}
\exp\!\Big(-\tfrac12(z-\mu_c)^\top\Sigma^{-1}(z-\mu_c)\Big),
\]
so $\sup_z \phi_c(z)=\phi_c(\mu_c)=(2\pi)^{-d/2}(\det\Sigma)^{-1/2}$.
Using $\exp(\cdot)\le 1$ and $\sum_c\pi_c=1$, we obtain for all $z$,
\[
p_\theta(z)=\sum_{c=1}^C \pi_c \phi_c(z)
\le
\sum_{c=1}^C \pi_c (2\pi)^{-d/2}(\det\Sigma)^{-1/2}
=
(2\pi)^{-d/2}(\det\Sigma)^{-1/2}.
\]
This proves \eqref{eq:app-linfty-lda}. Substituting into
Proposition~\ref{prop:app-overlap-kl} yields \eqref{eq:app-overlap-lda}.
\end{proof}

\begin{remark}[The failure mode matches covariance collapse]
\label{rem:app-overlap-failure}
The factor $(\det\Sigma)^{-1/2}$ in \eqref{eq:app-overlap-lda} can blow up as
$\det\Sigma\downarrow 0$. In that regime, minimizing $L_2$ may no longer imply
minimization of $L_1$ through this bound. This is precisely the covariance-collapse
pathology observed under unconstrained maximum-likelihood training.
\end{remark}

\subsection{Marginalization of KL divergence}
\label{app:kl-marginal}

We next record the standard monotonicity of KL under marginalization, which
connects $\mathrm{KL}(p(z)\|p_\theta(z))$ to the joint KL appearing in
\eqref{eq:app-L1}--\eqref{eq:app-L2}.

\begin{theorem}[Monotonicity of KL divergence under marginalization]
\label{thm:app-kl-marginal}
Let $Z\in\mathbb R^d$ be continuous and $Y\in\{1,\dots,C\}$ be discrete.
Assume $p(z,y)$ and $q(z,y)$ admit factorizations
\[
p(z,y)=p(z)\,p(y\mid z),
\qquad
q(z,y)=q(z)\,q(y\mid z),
\]
with conditional pmfs. If $p(z,y)\ll q(z,y)$, then
\[
\mathrm{KL}\big(p(z)\,\|\,q(z)\big)\le \mathrm{KL}\big(p(z,y)\,\|\,q(z,y)\big).
\]
\end{theorem}

\begin{proof}
By definition,
\[
\mathrm{KL}\big(p(z,y)\,\|\,q(z,y)\big)
=
\int_{\mathbb R^d}\sum_{y=1}^C p(z,y)\log\frac{p(z,y)}{q(z,y)}\,dz.
\]
Using the factorizations,
\[
\log\frac{p(z,y)}{q(z,y)}
=
\log\frac{p(z)}{q(z)}
+
\log\frac{p(y\mid z)}{q(y\mid z)}.
\]
Substituting and splitting terms,
\begin{align*}
\mathrm{KL}\big(p(z,y)\,\|\,q(z,y)\big)
&=
\int_{\mathbb R^d}\sum_{y} p(z,y)\log\frac{p(z)}{q(z)}\,dz
+
\int_{\mathbb R^d}\sum_{y} p(z,y)\log\frac{p(y\mid z)}{q(y\mid z)}\,dz.
\end{align*}
The first term simplifies since $\log\frac{p(z)}{q(z)}$ does not depend on $y$:
\[
\int_{\mathbb R^d}\sum_{y} p(z,y)\log\frac{p(z)}{q(z)}\,dz
=
\int_{\mathbb R^d} p(z)\log\frac{p(z)}{q(z)}\,dz
=
\mathrm{KL}\big(p(z)\,\|\,q(z)\big).
\]
The second term equals
\[
\mathbb E_{Z\sim p}\Big[\mathrm{KL}\big(p(\cdot\mid Z)\,\|\,q(\cdot\mid Z)\big)\Big]\ge 0,
\]
since KL between discrete distributions is nonnegative. Therefore,
\[
\mathrm{KL}\big(p(z,y)\,\|\,q(z,y)\big)\ge \mathrm{KL}\big(p(z)\,\|\,q(z)\big).
\]
\end{proof}

\subsection{When minimizing $L_2$ also decreases $L_1$}
\label{app:l1-vs-l2}

By Theorem~\ref{thm:app-kl-marginal}, for our model $q=p_\theta$,
\[
\mathrm{KL}\big(p(z)\,\|\,p_\theta(z)\big)
\le
\mathrm{KL}\big(p(z,y)\,\|\,p_\theta(z,y)\big)
\le
L_2(\theta).
\]
Combining this with Proposition~\ref{prop:app-overlap-lda}, we obtain
\begin{align}
\int p_\theta(z)^2\,dz
&\le
\int p(z)p_\theta(z)\,dz
+
(2\pi)^{-d/2}(\det\Sigma)^{-1/2}\sqrt{2\,\mathrm{KL}\big(p(z)\|p_\theta(z)\big)}
\nonumber\\
&\le
\int p(z)p_\theta(z)\,dz
+
(2\pi)^{-d/2}(\det\Sigma)^{-1/2}\sqrt{2\,L_2(\theta)}.
\label{eq:app-collision-upper}
\end{align}
Multiplying by $\lambda$ and adding the shared KL term gives the explicit comparison
\begin{multline}
\label{eq:app-L1-vs-L2}
L_1(\theta)
\le
L_2(\theta)
+
\lambda(2\pi)^{-d/2}(\det\Sigma)^{-1/2}\sqrt{2\,\mathrm{KL}\big(p(z)\|p_\theta(z)\big)}\\
\le
L_2(\theta)
+
\lambda(2\pi)^{-d/2}(\det\Sigma)^{-1/2}\sqrt{2\,L_2(\theta)}.
\end{multline}

\begin{remark}[The same caveat, now in LDA form]
\label{rem:app-caveat}
Equation \eqref{eq:app-L1-vs-L2} suggests that, as $L_2(\theta)$ decreases, the
intrinsic information potential-penalized objective $L_1(\theta)$ also decreases, \emph{provided}
$(\det\Sigma)^{-1/2}$ remains controlled. If $\det\Sigma\downarrow 0$, the factor
blows up and the bound becomes uninformative; minimizing $L_2$ may then fail to
track $L_1$ through this argument. This aligns with the degeneracy of unconstrained
likelihood training and motivates explicitly preventing covariance collapse (e.g.,
spherical parameterization and/or lower-bounding $\sigma^2$ in Section~4).
In the spherical case $\Sigma=\sigma^2 I_d$, the blow-up factor becomes $\sigma^{-d}$.
\end{remark}

\section{Additional calibration results on CIFAR-10 and Fashion-MNIST}
\label{app:calibration-additional}

We report additional confidence histograms and reliability diagrams (with ECE)
for CIFAR-10 and Fashion-MNIST, following the protocol of Section~\ref{sec:calibration}.
Across both datasets, the LDA head trained with DNLL is better calibrated than
the softmax baseline (ECE 3.08\% vs 7.73\% on CIFAR-10; 2.98\% vs 5.05\% on
Fashion-MNIST).

\begin{figure}[htbp]
  \centering
  \begin{subfigure}{0.48\linewidth}
    \centering
    \includegraphics[width=\linewidth]{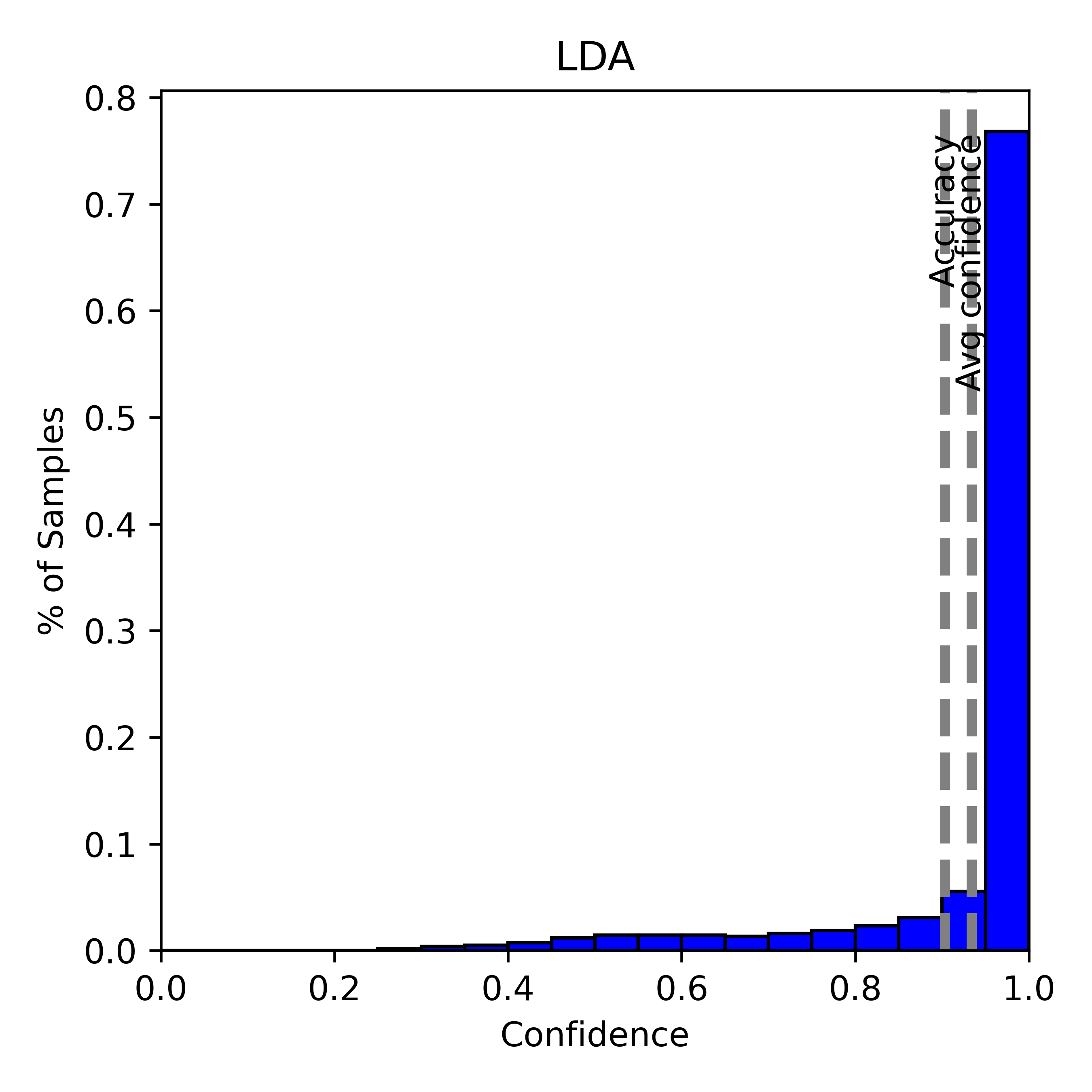}
    \caption{CIFAR-10: confidence histogram}
  \end{subfigure}\hfill
  \begin{subfigure}{0.48\linewidth}
    \centering
    \includegraphics[width=\linewidth]{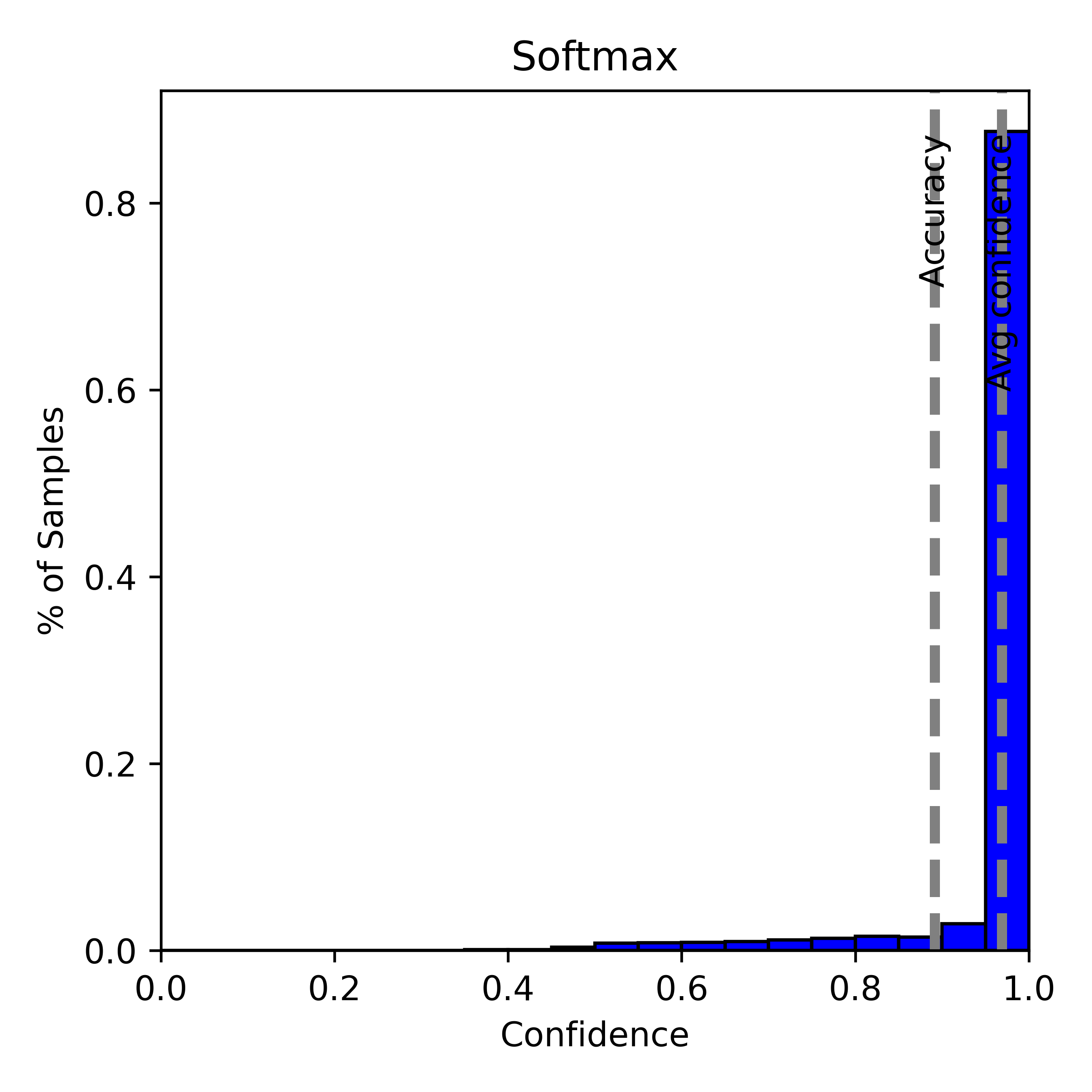}
    \caption{CIFAR-10: confidence histogram}
  \end{subfigure}

  \vspace{0.5em}

  \begin{subfigure}{0.48\linewidth}
    \centering
    \includegraphics[width=\linewidth]{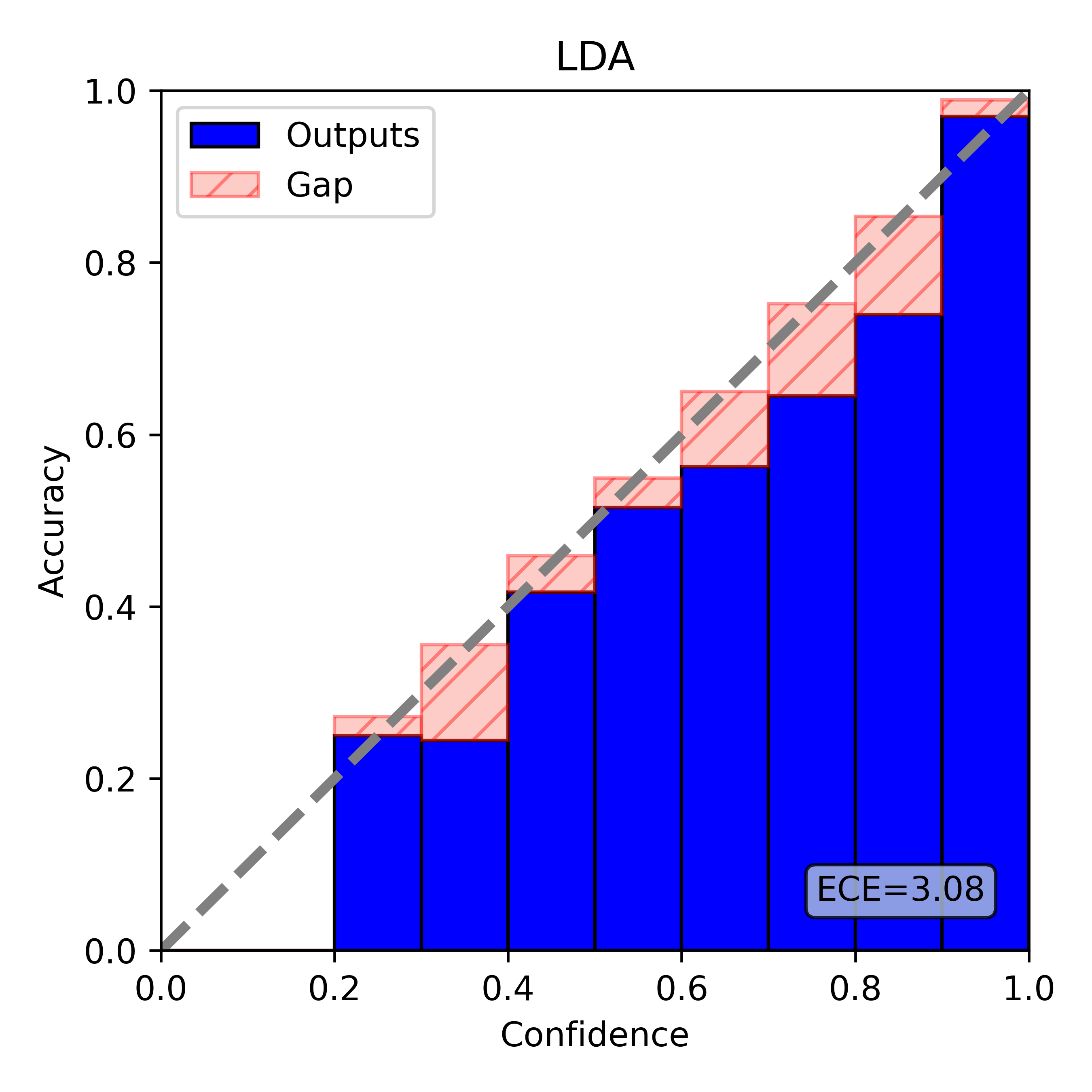}
    \caption{CIFAR-10: reliability diagram}
  \end{subfigure}\hfill
  \begin{subfigure}{0.48\linewidth}
    \centering
    \includegraphics[width=\linewidth]{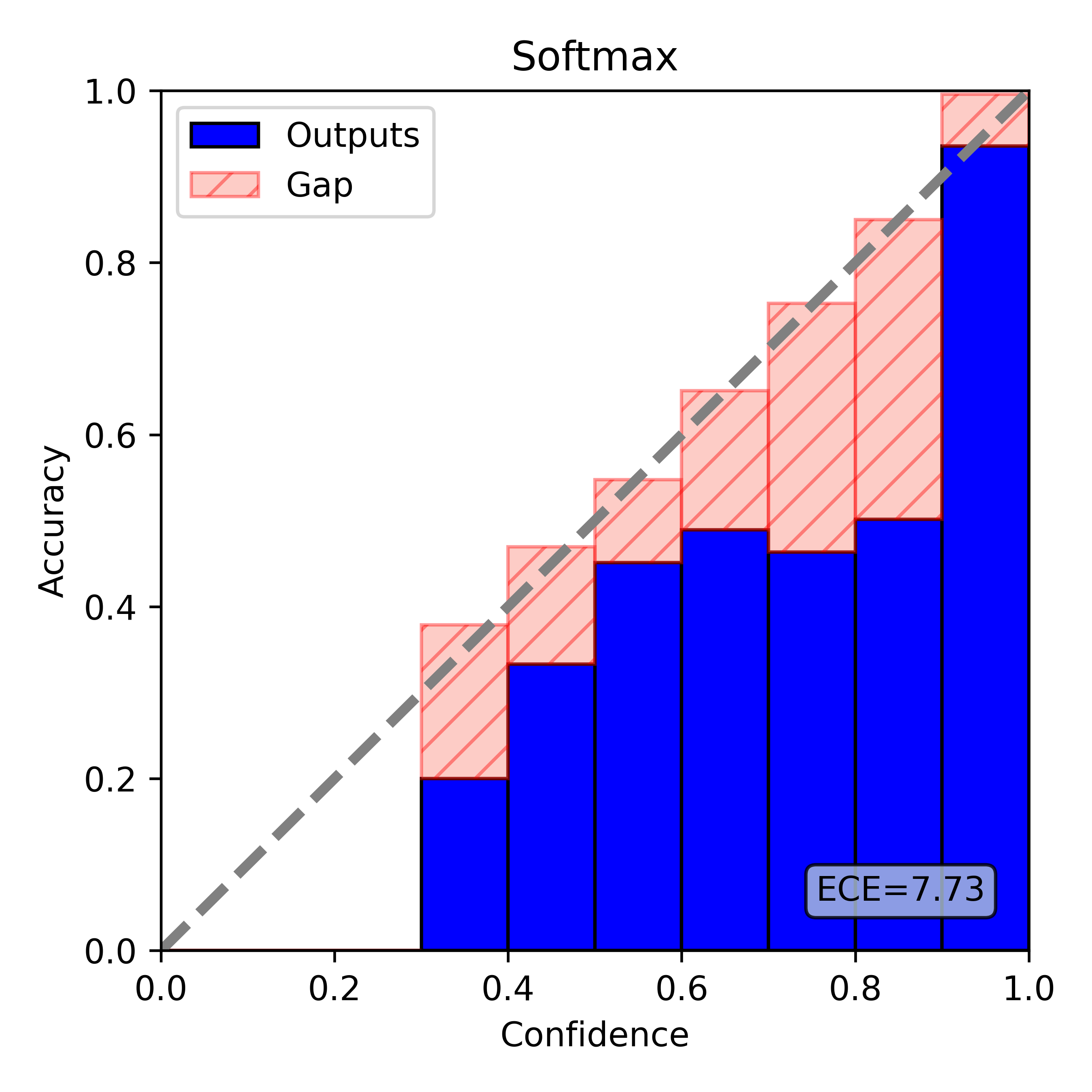}
    \caption{CIFAR-10: reliability diagram}
  \end{subfigure}
  \caption{Calibration and confidence behavior on CIFAR-10 (same format as Fig.~8).}
  \label{fig:calibration-cifar10}
\end{figure}

\begin{figure}[t]
  \centering
  \begin{subfigure}{0.48\linewidth}
    \centering
    \includegraphics[width=\linewidth]{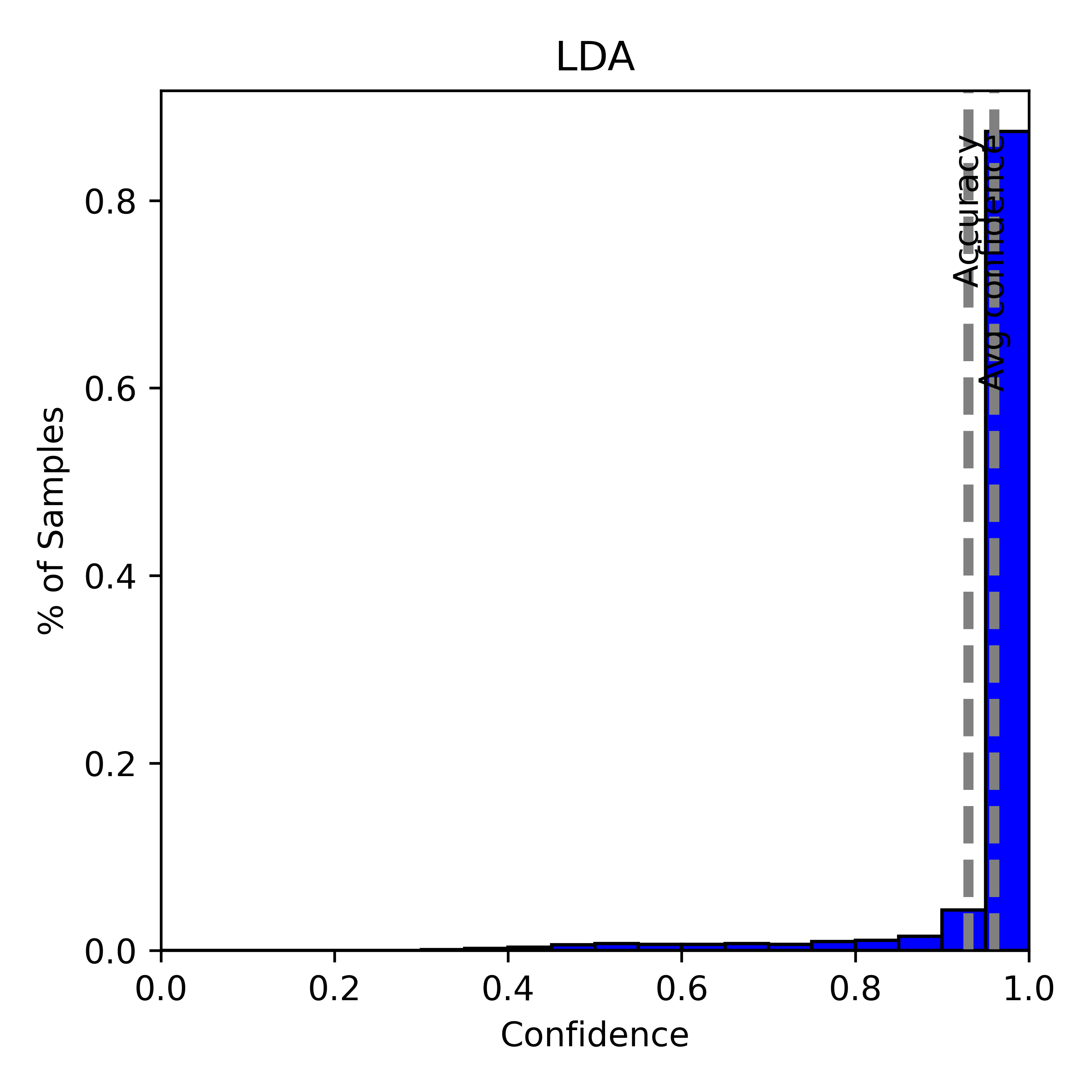}
    \caption{Fashion-MNIST: confidence histogram}
  \end{subfigure}\hfill
  \begin{subfigure}{0.48\linewidth}
    \centering
    \includegraphics[width=\linewidth]{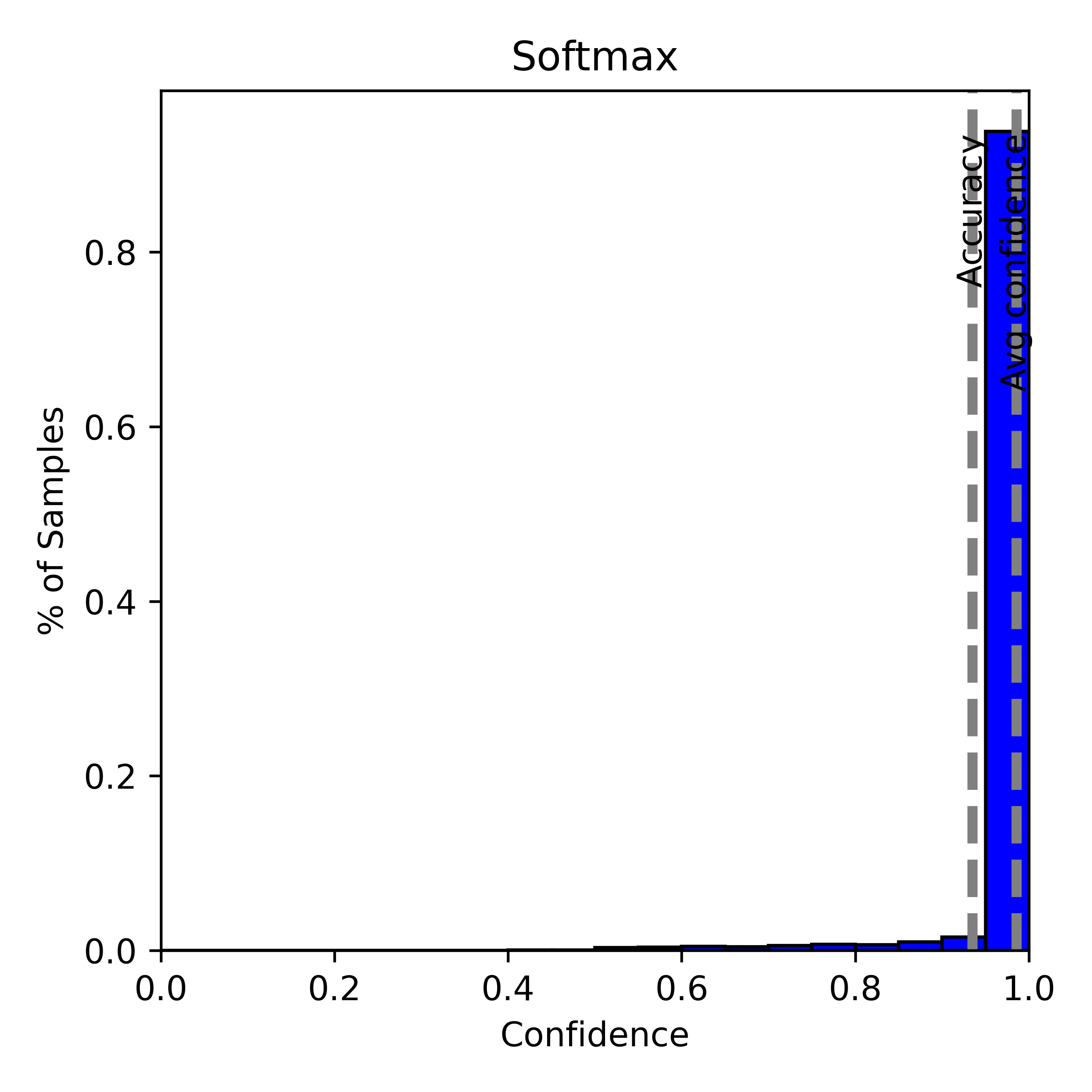}
    \caption{Fashion-MNIST: confidence histogram}
  \end{subfigure}

  \vspace{0.5em}

  \begin{subfigure}{0.48\linewidth}
    \centering
    \includegraphics[width=\linewidth]{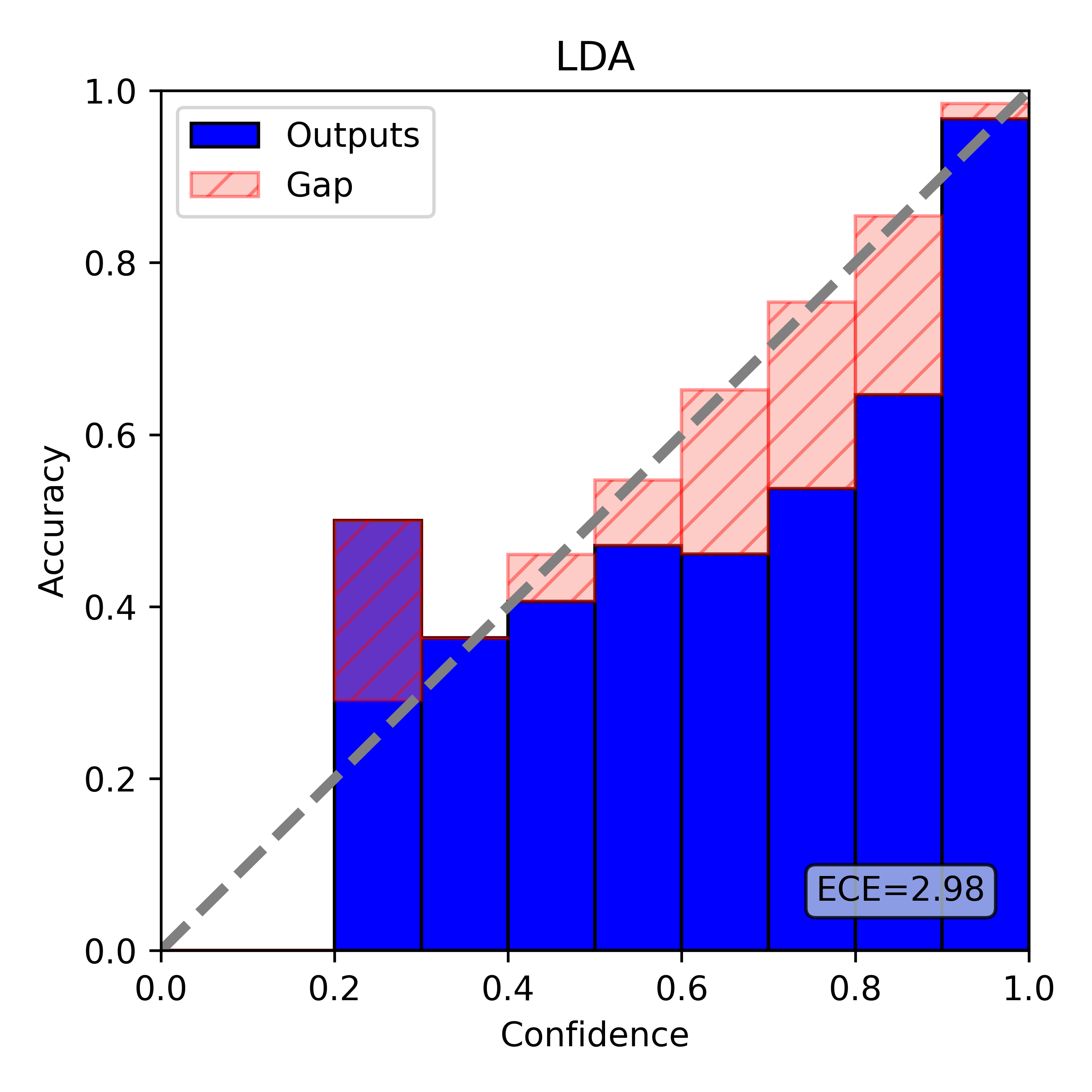}
    \caption{Fashion-MNIST: reliability diagram}
  \end{subfigure}\hfill
  \begin{subfigure}{0.48\linewidth}
    \centering
    \includegraphics[width=\linewidth]{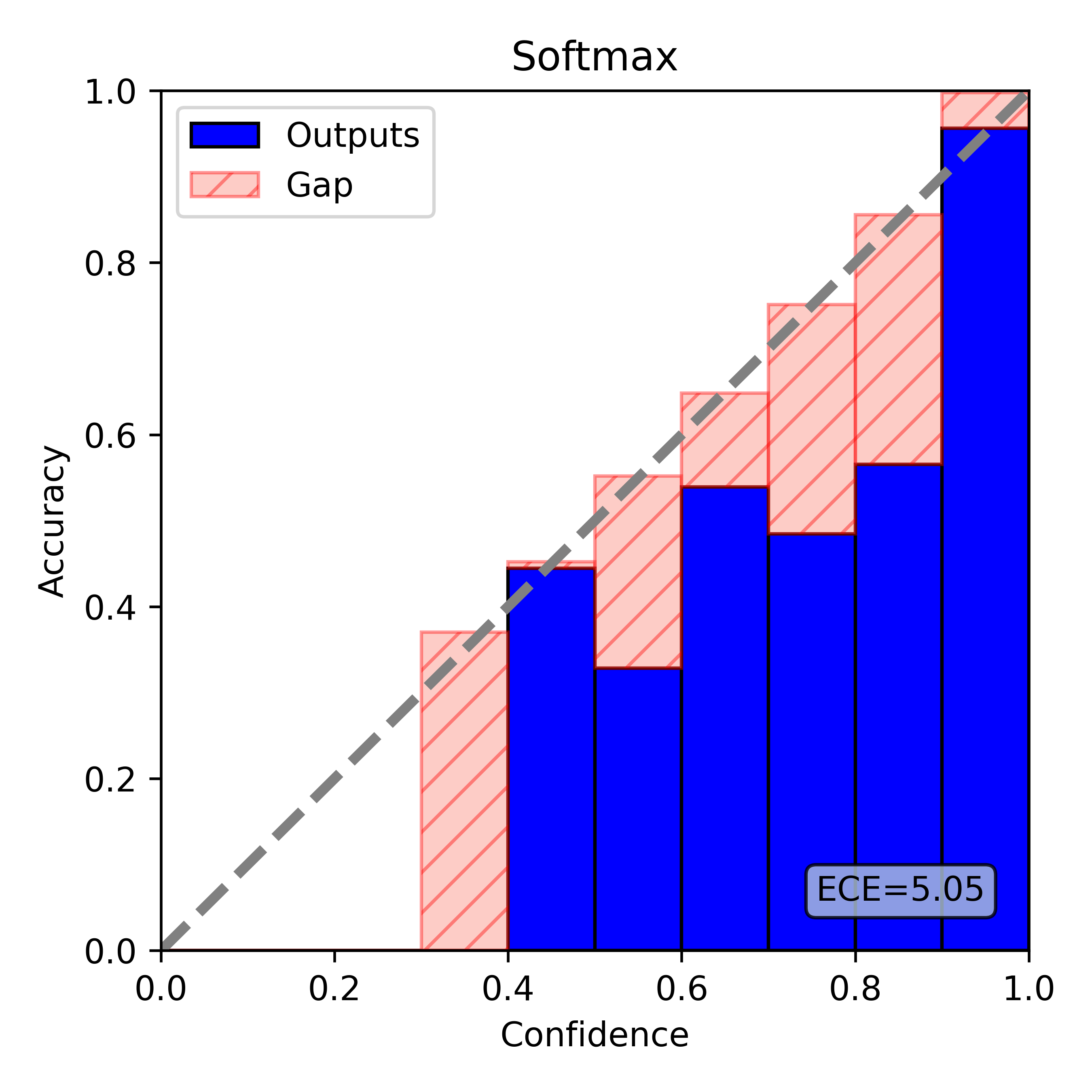}
    \caption{Fashion-MNIST: reliability diagram}
  \end{subfigure}
  \caption{Calibration and confidence behavior on Fashion-MNIST (same format as Fig.~\ref{fig:cifar100_calibration}).}
  \label{fig:calibration-fashion}
\end{figure}

\section{Additional analysis on sensitivity to the DNLL weight $\lambda$}\label{app:lambda}

\begin{figure}[htbp]
  \centering
  \begin{subfigure}[t]{0.32\textwidth}
    \centering
    \includegraphics[width=\linewidth]{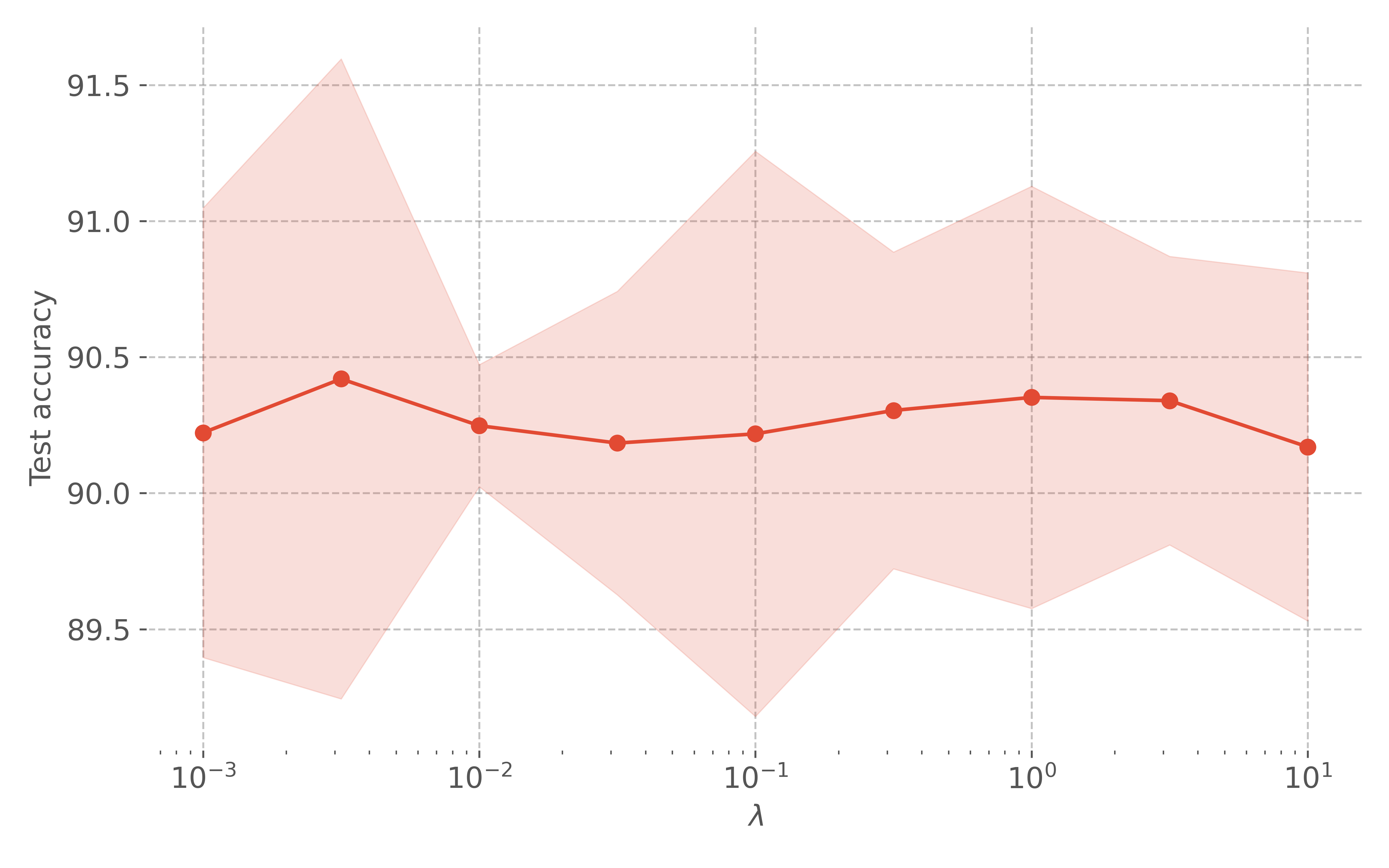}
    \caption{CIFAR-10: accuracy}
  \end{subfigure}\hfill
  \begin{subfigure}[t]{0.32\textwidth}
    \centering
    \includegraphics[width=\linewidth]{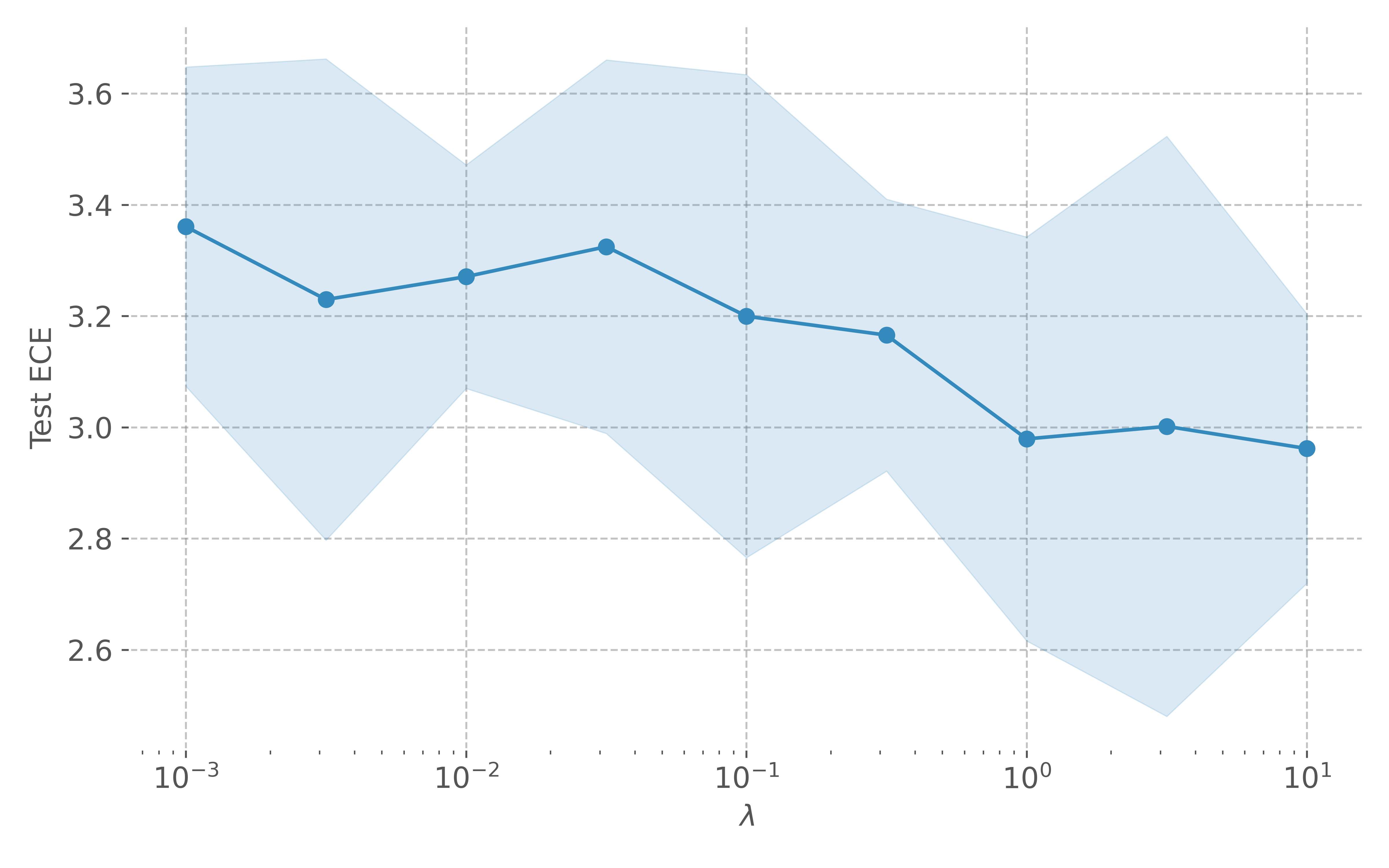}
    \caption{CIFAR-10: ECE}
  \end{subfigure}\hfill
  \begin{subfigure}[t]{0.32\textwidth}
    \centering
    \includegraphics[width=\linewidth]{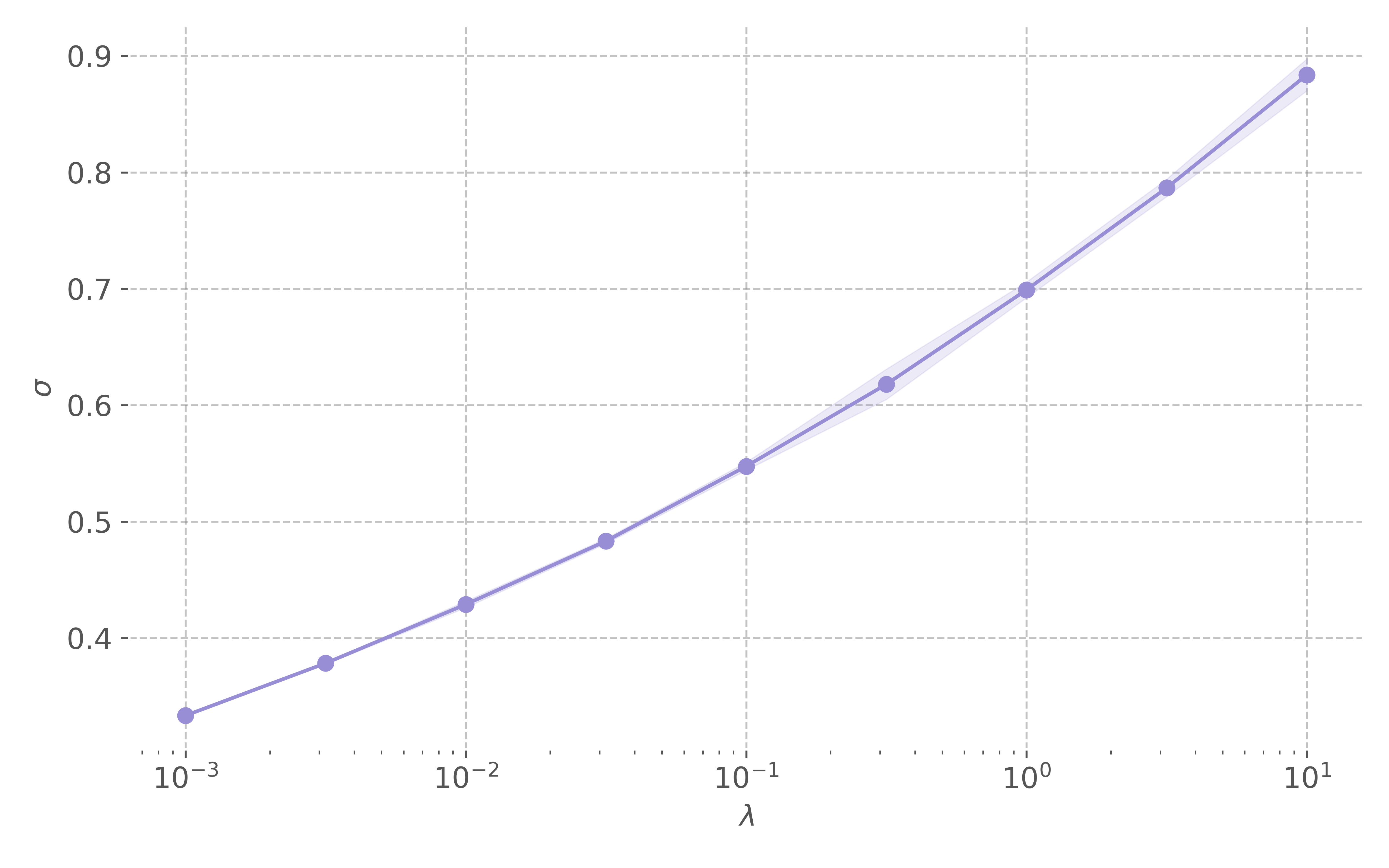}
    \caption{CIFAR-10: $\sigma$}
  \end{subfigure}

  \vspace{2mm}

  \begin{subfigure}[t]{0.32\textwidth}
    \centering
    \includegraphics[width=\linewidth]{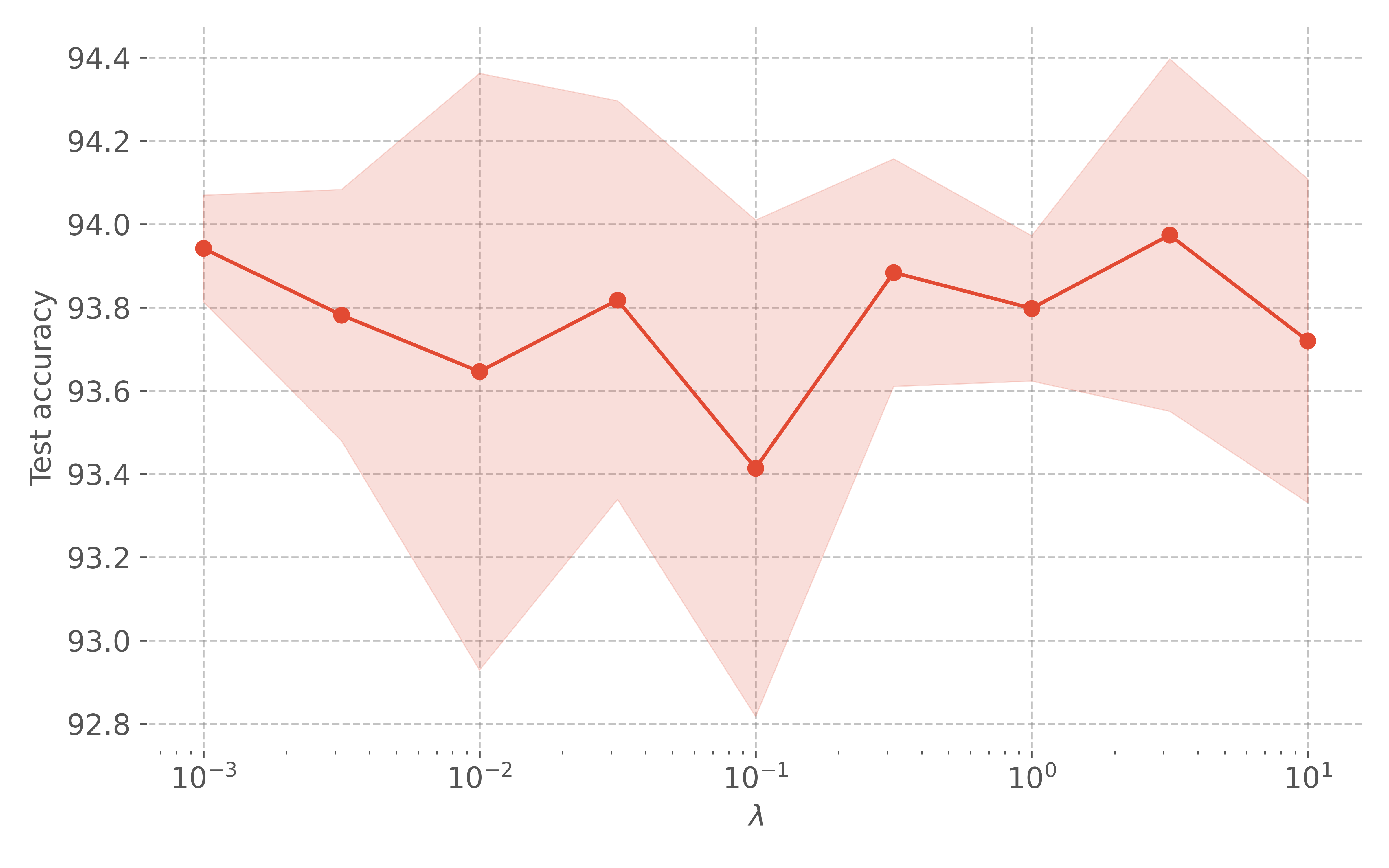}
    \caption{Fashion-MNIST: accuracy}
  \end{subfigure}\hfill
  \begin{subfigure}[t]{0.32\textwidth}
    \centering
    \includegraphics[width=\linewidth]{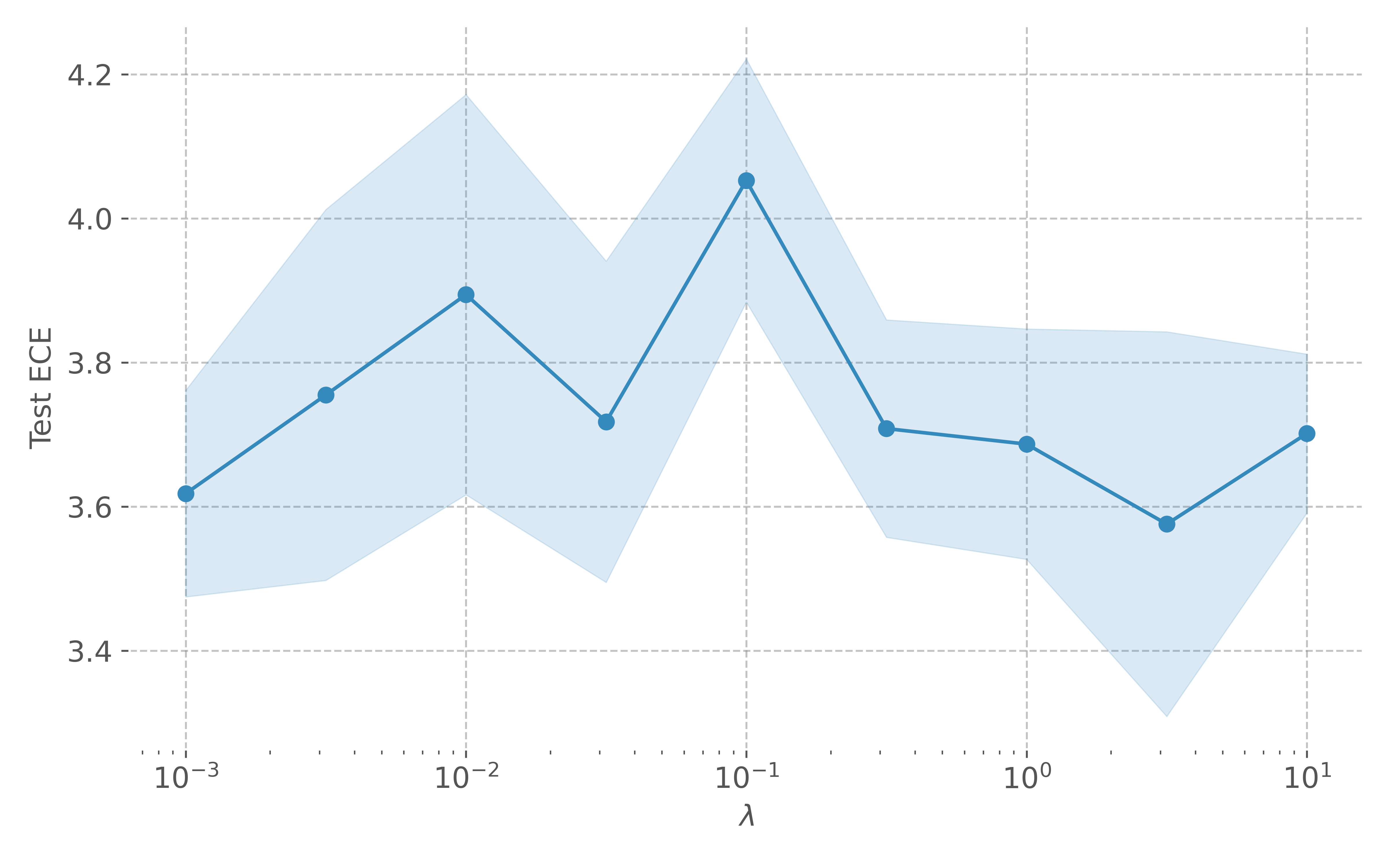}
    \caption{Fashion-MNIST: ECE}
  \end{subfigure}\hfill
  \begin{subfigure}[t]{0.32\textwidth}
    \centering
    \includegraphics[width=\linewidth]{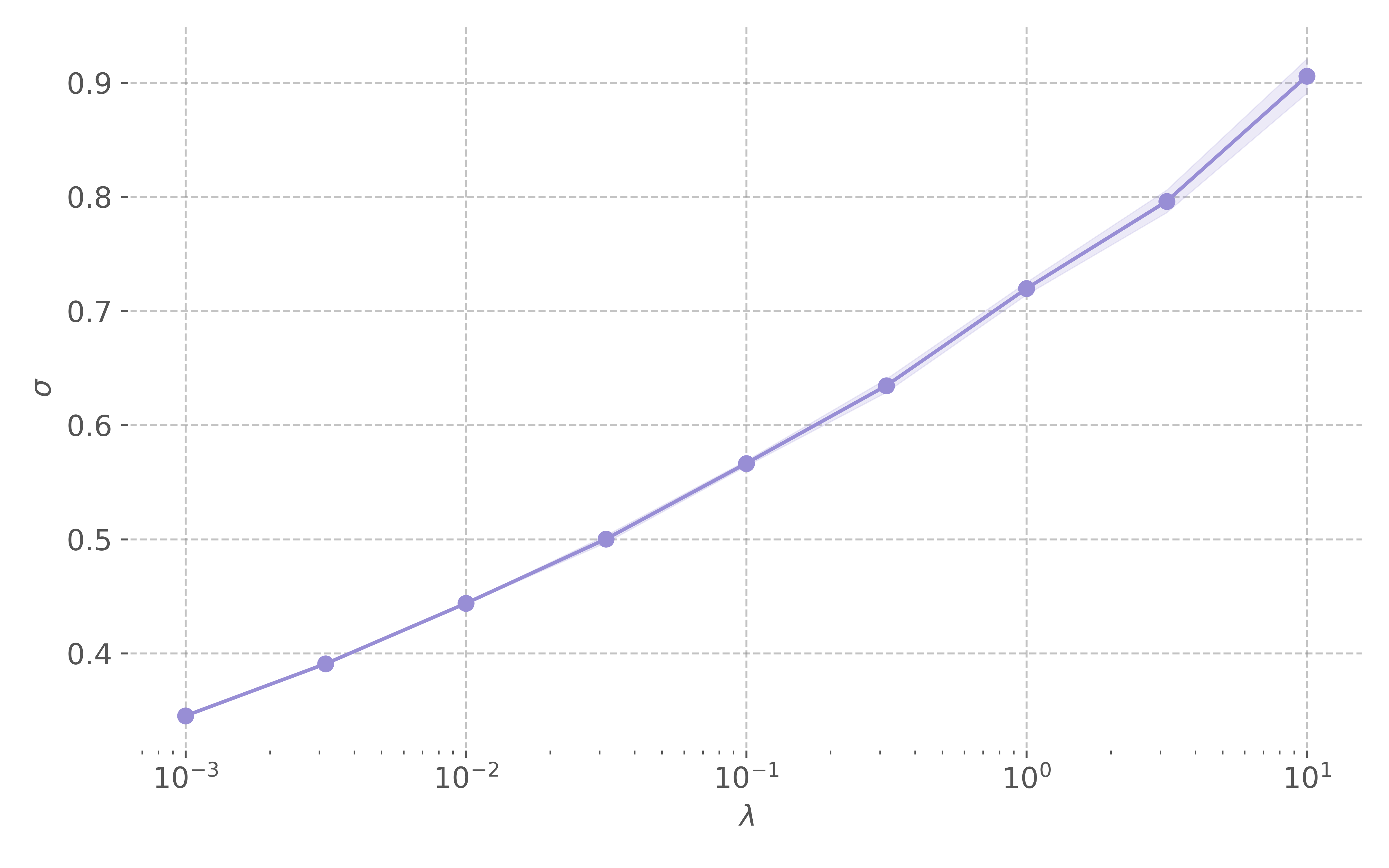}
    \caption{Fashion-MNIST: $\sigma$}
  \end{subfigure}

  \caption{Sensitivity of Deep LDA  to the regularization weight 
$\lambda$ on CIFAR-10 (top) and Fashion-MNIST (bottom). Left: test accuracy. Middle: test ECE. Right: learned spherical covariance scale 
$\sigma$. Markers denote the mean over runs; shaded regions show variability across seeds. Accuracy and ECE vary modestly across several orders of magnitude, while 
$\sigma$ changes systematically with 
$\lambda$, consistent with 
$\lambda$ acting primarily as a scale/temperature knob for the generative head.}
  \label{fig:lambda_sweep_c10_fmnist}
\end{figure}

We repeat the $\lambda$ sweep of Section~\ref{sec:real-data} on CIFAR-10 and Fashion-MNIST.
Figure~\ref{fig:lambda_sweep_c10_fmnist} shows that test accuracy and ECE are
stable across several orders of magnitude of $\lambda$, while the learned spherical
scale $\sigma$ varies monotonically with $\lambda$, supporting the interpretation
of $\lambda$ as a scale/temperature knob.

\end{document}